\documentclass{article}
\PassOptionsToPackage{numbers, compress}{natbib}
\usepackage[preprint]{neurips_2025}

\usepackage[utf8]{inputenc} % allow utf-8 input
\usepackage[T1]{fontenc}    % use 8-bit T1 fonts
\usepackage{hyperref}       % hyperlinks
\usepackage{url}            % simple URL typesetting
\usepackage{booktabs}       % professional-quality tables
\usepackage{amsfonts}       % blackboard math symbols
\usepackage{nicefrac}       % compact symbols for 1/2, etc.
\usepackage{microtype}      % microtypography
\usepackage{xcolor}         % colors
\usepackage{listings}
\usepackage{xcolor} % 用于颜色支持
\lstdefinestyle{python}{
    language=Python,
    basicstyle=\ttfamily\small,
    keywordstyle=\color{blue}\bfseries,
    commentstyle=\color{deepgreen},
    stringstyle=\color{red},
    numberstyle=\tiny\color{gray},
    showstringspaces=false,
    frame=single,
    breaklines=true,
    backgroundcolor=\color{lightgray!15}
}
\usepackage{times}
\usepackage{CJKutf8}
\usepackage{latexsym}

\usepackage{multirow}
\usepackage{wrapfig}
\usepackage{tikz}
\usepackage{capt-of}
\usepackage{graphicx}  %Required
\usepackage{pgfplots}
\pgfplotsset{compat=1.12}
\usepackage{amsmath}
\usepackage{multicol}
\usepackage{color}
\usepackage{mwe}
\usepackage{wrapfig}
\usepackage{colortbl,array}
\usepackage{xspace}
\usepackage{tikz}
\usetikzlibrary{tikzmark}
\makeatletter
\newcommand*\myfontsize{%
  \@setfontsize\myfontsize{7}{8}%
}
\makeatother
\newcommand{\mytextbox}[2]{\tikzmarknode[draw=#1,thick,inner sep=2pt]{test}{\myfontsize #2}}
\definecolor{myred}{rgb}{0.7, 0.3, 0.0}
\definecolor{myblue}{HTML}{054488}
\definecolor{mygreen}{HTML}{056b34}
\definecolor{myorange}{HTML}{ff8800}
\definecolor{mypurple}{HTML}{8400ff}
\definecolor{mypink}{HTML}{f7acb9}

\newcommand{\red}[1]{\mytextbox{myred}{\textbf{\textcolor{myred}{#1}}}}
\newcommand{\blue}[1]{\mytextbox{myblue}{\textbf{\textcolor{myblue}{#1}}}}
\newcommand{\green}[1]{\mytextbox{mygreen}{\textbf{\textcolor{mygreen}{#1}}}}

\newcommand{\purple}[1]{\mytextbox{mypurple}{\textbf{\textcolor{mypurple}{#1}}}}

\newcolumntype{R}[1]{>{\raggedleft\let\newline\\\arraybackslash\hspace{0pt}}m{#1}}

\usetikzlibrary{intersections}

\definecolor{darkgreen}{rgb}{0.0, 0.42, 0.24}
\usepackage{caption}
\usepackage{subcaption}
\usepackage{graphicx}
\usepackage{pifont}
\usepackage{titletoc}
\usepackage{amsfonts}
\usepackage{booktabs}
\usepackage{arydshln}
\usepackage{colortbl}
\usepackage{algorithm}
\usepackage[noend]{algpseudocode}
\usepackage{enumitem}
\usepackage{graphicx}
\usepackage{soul}
\usepackage{colortbl,array,xcolor}

\definecolor{citecolor}{HTML}{0051f4}
\definecolor{pink}{HTML}{ed008c}
\usepackage[utf8]{inputenc} % allow utf-8 input
\usepackage[T1]{fontenc}    % use 8-bit T1 fonts
\usepackage{hyperref}       % hyperlinks
\usepackage{url}            % simple URL typesetting
\usepackage{booktabs}       % professional-quality tables
\usepackage{amsfonts}       % blackboard math symbols
\usepackage{nicefrac}       % compact symbols for 1/2, etc.
\usepackage{microtype}      % microtypography
\usepackage{xcolor}         % colors

\usepackage{times}
\usepackage{CJKutf8}
\usepackage{latexsym}
\usepackage{multirow}
\usepackage{wrapfig}
\usepackage{tikz}
\usepackage{capt-of}
\usepackage{graphicx}  %Required
\usepackage{pgfplots}
\pgfplotsset{compat=1.12}
\usepackage{amsmath}
\usepackage{multicol}
\usepackage{color}
\usepackage{mwe}
\usepackage{wrapfig}
\usepackage{colortbl,array}
\usepackage{xspace}
\usepackage{tikz}
\usetikzlibrary{tikzmark}
\definecolor{myred}{rgb}{0.7, 0.3, 0.0}
\definecolor{myblue}{HTML}{054488}
\definecolor{mygreen}{HTML}{056b34}

\newcolumntype{R}[1]{>{\raggedleft\let\newline\\\arraybackslash\hspace{0pt}}m{#1}}

\usetikzlibrary{intersections}

\definecolor{darkgreen}{rgb}{0.0, 0.42, 0.24}
\usepackage{caption}
\usepackage{subcaption}
\usepackage{graphicx}
\usepackage{pifont}
\usepackage{titletoc}
\usepackage{amsfonts}
\usepackage{booktabs}
\usepackage{arydshln}
\usepackage{colortbl}
\usepackage{algorithm}
\usepackage[noend]{algpseudocode}
\usepackage{enumitem}
\usepackage{graphicx}
\usepackage{soul}
\usepackage{colortbl,array,xcolor}
\usepackage[listings,skins,breakable]{tcolorbox}

\definecolor{Ocean}{RGB}{129,194,250}

\definecolor{deepgreen}{RGB}{0, 70, 0}
\usepackage{cleveref}
\crefformat{section}{\S#2#1#3}
\crefformat{subsection}{\S#2#1#3}
\crefformat{subsubsection}{\S#2#1#3}
\crefrangeformat{section}{\S\S#3#1#4 to~#5#2#6}
\crefmultiformat{section}{\S\S#2#1#3}{ and~#2#1#3}{, #2#1#3}{ and~#2#1#3}
\crefmultiformat{subsection}{\S\S#2#1#3}{ and~#2#1#3}{, #2#1#3}{ and~#2#1#3}
\Crefformat{figure}{#2Fig.~#1#3}
\Crefmultiformat{figure}{Figs.~#2#1#3}{ and~#2#1#3}{, #2#1#3}{ and~#2#1#3}
\Crefformat{table}{#2Tab.~#1#3}
\Crefmultiformat{table}{Tabs.~#2#1#3}{ and~#2#1#3}{, #2#1#3}{ and~#2#1#3}
\Crefformat{appendix}{Appx.~\S#2#1#3}
\crefmultiformat{appendix}{Appx.~\S#2#1#3}{ and~#2#1#3}{, #2#1#3}{ and~#2#1#3}
\crefformat{algorithm}{Alg.~#2#1#3}
\Crefformat{equation}{Eq.~#2#1#3}

\usepackage{xcolor}
\definecolor{deepred}{RGB}{180,0,0}

\definecolor{citecolor}{HTML}{001dc6}
\definecolor{pink}{HTML}{ed008c}
\hypersetup{
    colorlinks,
    linkcolor=citecolor,
    urlcolor=pink,
    citecolor=citecolor
}

\title{Tool-Star: Empowering LLM-Brained Multi-Tool Reasoner via Reinforcement Learning}
\author{
Guanting Dong$^1$, Yifei Chen$^1$, Xiaoxi Li$^1$, Jiajie Jin$^1$, Hongjin Qian$^2$, Yutao Zhu$^1$  \\
\textbf{Hangyu Mao$^3$, Guorui Zhou$^3$, Zhicheng Dou$^1$\thanks{Corresponding author.}, ~ Ji-Rong Wen$^1$} \\
$^1$Renmin University of China ~~ $^2$BAAI~~ $^3$Kuaishou Technology\\
\texttt{\{dongguanting, dou\}@ruc.edu.cn}}

\begin{document}
\begin{CJK}{UTF8}{gbsn}

\maketitle

\begin{abstract}

Recently, large language models (LLMs) have shown remarkable reasoning capabilities via large-scale reinforcement learning (RL). However, leveraging the RL algorithm to empower effective multi-tool collaborative reasoning in LLMs remains an open challenge. In this paper, we introduce Tool-Star, an RL-based framework designed to empower LLMs to autonomously invoke multiple external tools during stepwise reasoning. Tool-Star integrates six types of tools and incorporates systematic designs in both data synthesis and training. To address the scarcity of tool-use data, we propose a general tool-integrated reasoning data synthesis pipeline, which combines tool-integrated prompting with hint-based sampling to automatically and scalably generate tool-use trajectories. A subsequent quality normalization and difficulty-aware classification process filters out low-quality samples and organizes the dataset from easy to hard. Furthermore, we propose a two-stage training framework to enhance multi-tool collaborative reasoning by: (1) cold-start fine-tuning, which guides LLMs to explore reasoning patterns via tool-invocation feedback; and (2) a multi-tool self-critic RL algorithm with hierarchical reward design, which reinforces reward understanding and promotes effective tool collaboration. Experimental analyses on over 10 challenging reasoning benchmarks highlight the effectiveness and efficiency of Tool-Star. The code is available at \url{https://github.com/dongguanting/Tool-Star}.

\end{abstract}

\begin{figure}[H]
\centering
\vspace{-1em}
\includegraphics[width=0.95\linewidth]{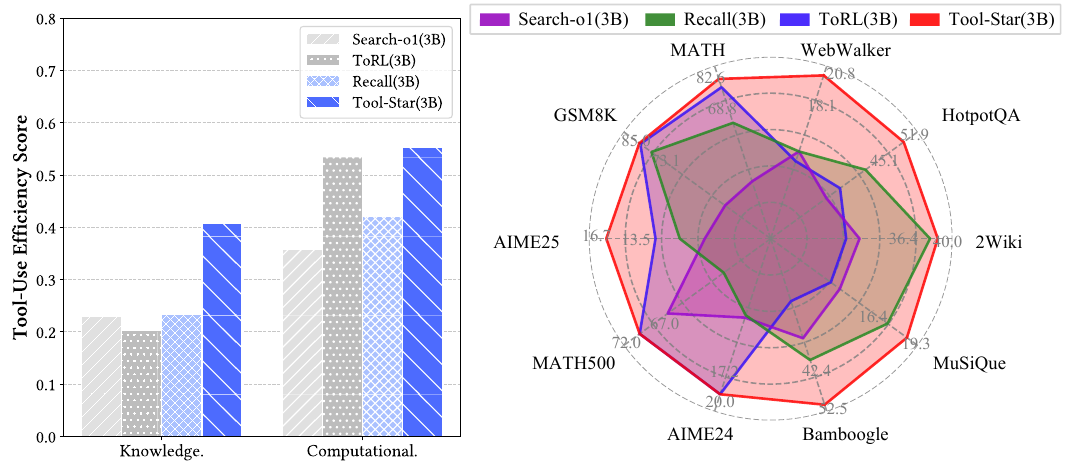}
% % caption
% \vspace{-0.5em}
\caption{
Overview result comparison of Tool-Star and other TIR baselines. \textbf{Left}: Average tool-use accuracy (TE) on knowledge-intensive and computational reasoning datasets. \textbf{Right}: The performance overview on 10 challenging reasoning datasets. }
\label{fig:intro}
\end{figure}

\section{Introduction}

Recent advances in large language models (LLMs) have demonstrated remarkable reasoning capabilities empowered by large-scale reinforcement learning (RL)~\citep{team2025kimi,reasoning_survey,reasoning_survey2,reasoning_survey3,simplerlzoo}. Advanced models exemplified by Deepseek R1~\citep{deepseek-r1} and OpenAI o1~\citep{2409_openai_o1} exhibit diverse emergent behaviors in Chain-of-Thought (CoT)
reasoning, including deep thinking, self-reflection, thereby improving problem-solving performance on complex reasoning tasks~\citep{2410_o1_journey_part1,2412_min_imitate}. However, real-world reasoning scenarios often require models to integrate multiple capabilities through interaction with the environment, such as deep information seeking, long-term knowledge memory, and precise computation. To achieve this, Tool-Integrated Reasoning (TIR)~\citep{tora,start,ragen,code_reasoning_survey,art} approaches encourage LLMs to interact with external tools during the reasoning process, substantially expanding the paradigm from language-only reasoning to a broader scope of applications.

Current foundational work on TIR primarily focuses on distilling tool-use trajectories from strong models and applying supervised fine-tuning (SFT) to guide weaker models via imitation learning~\citep{tora,dotamath,siam,rag-star}. As large reasoning models exhibit emergent capabilities, subsequent efforts aim to enhance long chain-of-thought reasoning in LLMs through tool-integrated prompting design~\citep{pot,chainofcode,searcho1,oxford_agent,ircot}. However, these approaches rarely enable LLMs to autonomously discover effective reasoning patterns of tool usage. To address this limitation, recent studies introduce outcome-based rewards~\citep{orm,deepseekmath}, extending reinforcement learning to the TIR domain~\citep{r1searcher,searchr1,chen2025research,retool,torl,wang2025otc,qian2025toolrlrewardtoollearning}. 
While RL-based methods encourage exploration of efficient tool-use behaviors, they primarily focus on interacting with the individual tool.  Notably, real-world reasoning tasks often require LLMs to possess both accurate computation and dynamic information-seeking capabilities.\footnote{\url{https://openai.com/index/introducing-o3-and-o4-mini/}} This necessitates deeper integration of feedback from multiple tools, particularly search engines and code generators. The lack of systematic investigation into multi-tool collaborative reasoning presents a major bottleneck for practical TIR deployment. In this work, we aim to bridge this gap by addressing the following research questions:
\begin{enumerate}[leftmargin=1em]
\item \textbf{Rationality and efficiency of tool usage.} How to enable the model to perform efficient tool-augmented reasoning while balancing the cost of tool invocation.
\item \textbf{Multi-tool collaborative reasoning.} How to effectively enable models to integrate the functionalities of multiple tools throughout the reasoning process.
\end{enumerate}
In this paper, we propose \textbf{Tool-Star}, a reinforcement learning-based framework designed to empower LLMs to autonomously invoke multiple external tools during stepwise reasoning. Specifically, Tool-Star integrates six types of tools into the reasoning process (three for training and three for inference-time optimization) and incorporates systematic designs in both data synthesis and training algorithms.

To address the scarcity of tool-use data, we first design a general \textbf{Tool-Integrated Reasoning Data Synthesis Pipeline} that combines tool-integrated prompting with hint-based sampling to automatically generate large-scale tool-use trajectories. We then introduce a quality normalization and difficulty-aware classification process to effectively filter out unreasonable tool-use samples and partition the data in a curriculum-like manner from easy to hard~\citep{DBLP:journals/ijcv/SovianyIRS22,how_abilities,progressive}. Leveraging this pipeline, we construct high-quality datasets for both cold-start fine-tuning and reinforcement learning in a staged manner, laying a solid foundation for subsequent TIR training.

To incentivize the model’s capability for multi-tool collaboration, we propose a two-stage TIR training framework that progressively aligns reasoning abilities in an easy-to-hard manner. (1) In the first stage, we introduce a \textbf{Cold-Start Supervised Fine-Tuning} strategy, allowing LLMs to initially explore reasoning patterns with feedback from tool invoking. (2) In the second stage, we develop a \textbf{Multi-Tool Self-Critic Reinforcement Learning Algorithm}. Unlike prior RL approaches that focus on single-tool usage, our method employs a hierarchical reward mechanism that not only evaluates answer correctness and tool-use format but also assigns additional rewards for effective multi-tool collaboration. To further improve the model’s understanding of this complex reward structure, we interleave a self-critic reward fine-tuning phase within the standard RL process, facilitating the internalization of reward principles.

To comprehensively evaluate the capability of Tool-Star, we design experiments across more than 10 challenging computational reasoning tasks (e.g., AIME24 and MATH500~\cite{math500}) and knowledge-intensive reasoning tasks (e.g., WebWalker~\cite{2501_WebWalker} and HotpotQA~\cite{hotpotqa}). As illustrated in Figure~\ref{fig:intro}, Tool-Star demonstrates strong overall reasoning performance while ensuring both efficiency and reliability in tool usage. In summary, our main contributions are as follows:
\begin{itemize}[leftmargin=1em]

\item We propose Tool-Star, a two-stage training framework to enhance multi-tool collaborative reasoning in LLMs: (1) Cold-Start fine-tuning enables LLMs to explore reasoning patterns guided by tool-invocation feedback; (2) Multi-Tool self-critic RL algorithm with a hierarchical reward design reinforces the LLM’s grasp of reward principles and promotes effective multi-tool collaboration.

\item To address the scarcity of tool-use data, we propose a general TIR data synthesis pipeline that combines tool-integrated prompting with hint-based sampling to automatically and scalably generate tool-use trajectories. We further introduce a quality normalization and difficulty-aware classification process to filter out unreasonable samples and organize the dataset from easy to hard.

\item Experiments on more than 10 challenging reasoning datasets validate the effectiveness of Tool-Star. Further quantitative analyses demonstrate its high efficiency, offering insights for incentivizing multi-tool collaborative reasoning.
\end{itemize}

\section{Related Work}

\paragraph{Tool-Integrated Reasoning.} Tool-Integrated Reasoning (TIR) has emerged as a promising paradigm for enhancing LLM reasoning by enabling autonomous invocation of external tools. Existing TIR approaches can be broadly categorized into three streams: \textbf{(1) Prompting-based methods} guide models to use tools via carefully crafted prompts without additional training~\citep{pot,chainofcode,OctoTools,searcho1,oxford_agent,ierc,ircot}. While easy to implement, they often suffer from instability and limited accuracy in tool usage. \textbf{(2) SFT-based methods} apply supervised fine-tuning to teach weaker models tool use by distilling trajectories from stronger models~\citep{tora,dotamath,siam,rag-star,retrollm,autoif,dparag,start}. Though effective, their performance is constrained by the quality of demonstrations and limited generalization beyond seen examples. \textbf{(3) RL-based methods} optimize tool-use behavior via outcome-driven rewards~\citep{r1searcher,zerosearch,searchr1,chen2025research,retool,torl,webthinker,2505_ARTIST}, enabling models to autonomously discover effective tool-use strategies through exploration. This often leads to more robust and adaptive tool utilization than imitation-based approaches. Despite these advances, most work focuses on single-tool settings. Multi-tool collaborative reasoning—requiring coordination across heterogeneous tools (e.g., search engines, code interpreters)—remains underexplored. While systems like OTC~\citep{wang2025otc} and ToolRL~\citep{qian2025toolrl} have made initial progress, there is still a significant gap in developing specialized RL algorithms and reward mechanisms tailored for multi-tool collaboration.

\textbf{Reinforcement Learning for Reasoning.} Recent advances in reinforcement learning (RL) have significantly enhanced reasoning capabilities in language models. Large-scale RL implementations such as DeepSeek-R1~\citep{deepseek-r1}, OpenAI-o1~\citep{2409_openai_o1}, and QwQ~\citep{qwen_qwq} have demonstrated the effectiveness of RL for improving reasoning performance. Concurrently, several studies have explored achieving emergent reasoning capabilities with minimal resources, either through high-quality but small-scale datasets or smaller parameter models~\citep{simplerlzoo,limo,limr}. Notable examples include SimpleRL-Zoo~\citep{simplerlzoo}, LIMO~\citep{limo}, LIMR~\citep{limr}, and Qwen's mathematical reasoning approach~\citep{qwen2.5math}, with additional insights from works examining RL effectiveness in both small~\citep{rl-reasoning-small-LLMs} and large models with minimal examples~\citep{rl-one-example}.

A series of research has focused on analyzing RL training strategies and reward function design for reasoning tasks, providing practical insights for broader applications. Key contributions include PRIME~\citep{prime}, which explores process reinforcement through implicit rewards, empirical studies on improving R1-like reasoning models~\citep{2503_chenzp_empirial_r1, rl-kr-survey}, and DAPO's scalable RL system~\citep{DAPO}. Another research direction has concentrated on optimizing reasoning efficiency through specialized RL algorithms, as demonstrated in works on concise reasoning~\citep{concise-reasoning, wemath,stop-overthinking}. More recently, researchers have begun integrating external tools to enhance reasoning accuracy, with notable examples including ToRL~\citep{torl}, ReTool~\citep{retool}, ToolRL~\citep{qian2025toolrlrewardtoollearning}, and OTC~\citep{wang2025otc}. However, exploration of multi-tool collaborative reasoning remains relatively limited, highlighting a significant gap in current research.

\section{Methodology}
Tool-Star is an RL-based framework that empowers LLMs to autonomously invoke multiple external tools during reasoning. As shown in Figure~\ref{fig:method1} and Figure~\ref{fig:method2}, Tool-Star includes two key components: \textbf{(1)} A tool-integrated data synthesis pipeline that generates large-scale, high-quality reasoning trajectories with reasonable tool usage~(\Cref{sec:tool_data_synthesis}); \textbf{(2)} A two-stage training paradigm that facilitates multi-tool collaborative reasoning, comprising cold-start fine-tuning and a multi-tool self-critic RL algorithm~(\Cref{sec:tool_star_training}). We delve into the specifics of our approach as follows.

\textbf{Problem Formulation.} \textit{Multi-tool Collaborative Reasoning} aims to enable LLMs to perform multi-step reasoning through interaction and collaboration with multiple external tools. Specifically, given a task query $q$ and an external tool set $\mathcal{T}$, Tool-Star autonomously invokes tools during the generation of the reasoning chain $\mathcal{R}^{\text{c}}$, guided by the tool-integrated instruction $I^{\mathcal{T}}$. The real-time tool-call feedback $F^{\mathcal{T}}$ is dynamically concatenated into the reasoning chain to facilitate ongoing inference until the final output $y$ is produced. This process can be modeled as:
\begin{equation}
\small
P_{\theta}(\mathcal{R}^{\text{c}}, y \mid I_{\mathcal{T}}, q, \mathcal{T}) = \underbrace{\prod\nolimits_{t=1}^{T_{c}} P_{\theta}(\mathcal{R}^{\text{c}}_t \mid \mathcal{R}^{\text{c}}_{<t}, I^{\mathcal{T}}, q, \{F^{\mathcal{T}}\}_{< t})}_{\text{Multi-Tool Integrated Reasoning}} \cdot \underbrace{\prod\nolimits_{t=1}^{T_y} P_{\theta}(y_t \mid y_{<t}, \mathcal{R}_{\text{c}}, I^{\mathcal{T}}, q)}_{\text{Answer Generation}},
\label{equ:reason_with_tools}
\end{equation}
where $T_{c}$ denotes the number of tokens in the reasoning chain $\mathcal{R}^{\text{c}}$, $\mathcal{R}^{\text{c}}_t$ is the token at position $t$, and $\mathcal{R}^{\text{c}}_{<t}$ represents all tokens generated before $t$. $\{F^{\mathcal{T}}\}_{ < t}$ denotes the feedback of all tool calls prior to $t$. $T_y$ is the length of the Answer $y$, with $y_t$ as the token at position $t$.

\textbf{Tool Design.} In this work, we design six tools to enhance the reasoning capabilities of LLMs within the Tool-Star framework. During the two-stage training process, we introduce three core tools that enable LLMs to autonomously invoke external functionalities during reasoning:

\begin{itemize}[leftmargin=1em]
\item \textbf{Search Engine:} Executes search queries to retrieve relevant information, supporting both local and web-based search modes. 

\item \textbf{Web Browser Agent: }Parses web search results by visiting URLs, extracting relevant content, and summarizing key information in response to the query. 

\item \textbf{Code Interpreter:} Executes code snippets generated by the LLM in a sandbox environment, returning either the execution results or error messages based on code correctness.
\end{itemize}

During inference, we introduce three Inference-time tools to optimize the reliability of TIR process: 
\begin{itemize}[leftmargin=1em]
\item \textbf{Code Debugger:} Automatically corrects LLM-generated erroneous code by leveraging the original code and compiler error messages to guide revisions. 

\item \textbf{Tool-Use Backtracer:} Locates and rolls back to the reasoning step preceding a failed tool invocation, enabling the model to resume and revise its reasoning path. 

\item \textbf{Reasoning Chain Refiner:} When the output exceeds the maximum length, refiner prunes and optimizes redundant steps in the reasoning process, replacing original reasoning chain with a more concise and coherent version to complete the task.
\end{itemize}

\subsection{Tool-Integrated Reasoning Data Synthesis}
\label{sec:tool_data_synthesis}
In this section, we present our tool-integrated reasoning data synthesis pipeline, aiming to enable automated and scalable construction of high-quality tool-use datasets (Figure~\ref{fig:method1}).

\subsubsection{Step-1: Data Collection and Sampling} 
To balance data scale and diversity while ensuring accessibility, we curate a high-quality training set from open-source knowledge-based and computational reasoning datasets, comprising approximately 90K text-based reasoning data ($D_{\text{text}}$) and 1K existing TIR datasets ($D_{\text{tool}}$), as summarized in Table~\ref{tab:data_source}. To further expand  TIR trajectories, we introduce two complementary sampling strategies:
% Detailed dataset statistics are provided in the appendix.

\textbf{(1) TIR Prompting-based Sampling.} We design a tool-integrated prompt $I^T$ to guide the LLM $P_{\theta}$ in sampling responses for queries in $D_{\text{text}}$. Following the Eq.(\ref{equ:reason_with_tools}), LLM decodes tool-invoking requests within special tokens (e.g., \green{<search>}... \green{</search>}, \blue{<python>}... \blue{</python>}) during trajectory generation. We then automatically parse and extract these requests, invoke external tools to obtain feedback $F$, and insert the tool feedback—enclosed within \red{<result>} and \red{</result>} tags—back into the reasoning chain as additional context for subsequent generation steps. This process iterates until either (1) the maximum number of tool calls or length is reached, or (2) the model generates a final answer, enclosed by designated tokens \red{<answer>} and \red{</answer>}. After inference on the entire $D_{\text{text}}$, we filter for correct samples to obtain the dataset $D_{\text{tool}}^{P}$.

\begin{figure*}[!t]
\centering
\includegraphics[width=0.985\linewidth]{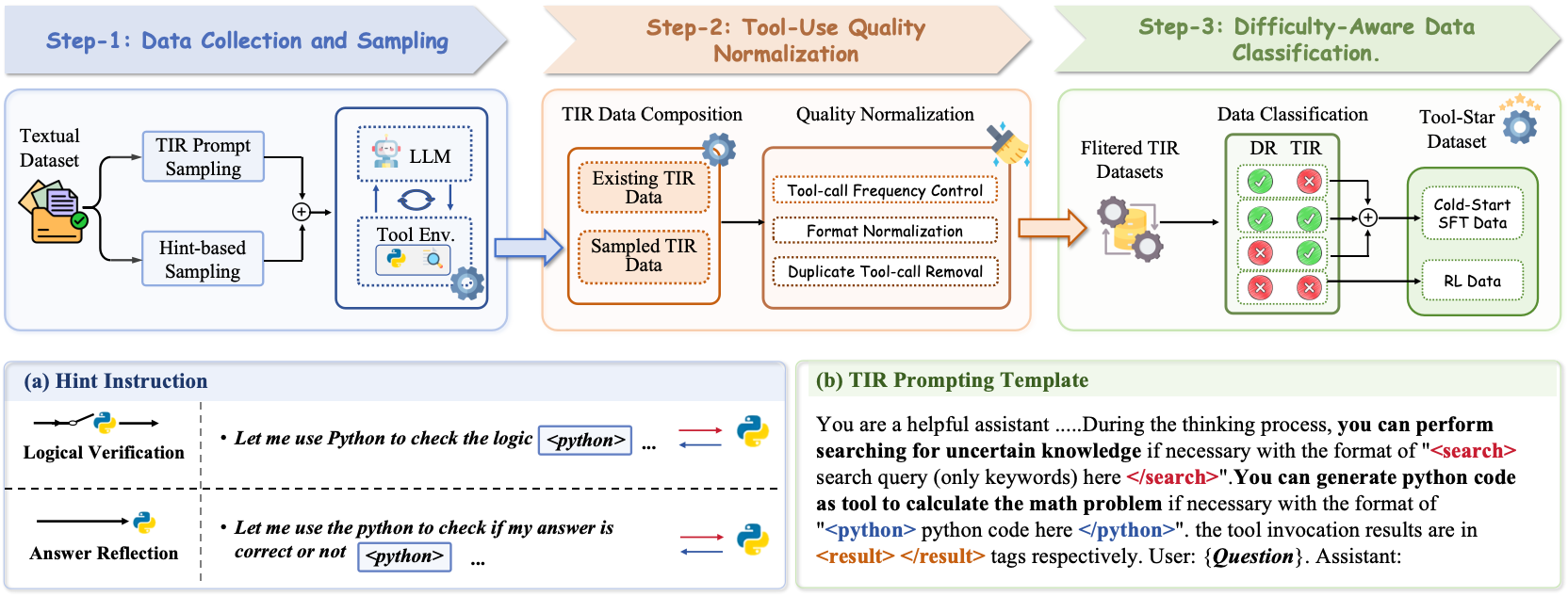}

\caption{The overview of 3-step tool-integrated reasoning data synthesis pipeline.}
\label{fig:method1} 
% \vspace{-1em}
\end{figure*}

\textbf{(2) Hint-based Sampling.} To further diversify tool invoking patterns, we employ a hint-based method~\citep{star} that inserts hint tool-call tokens into language-only reasoning trajectories. We first prompt the LLM to perform language-only reasoning on queries from $D_{\text{text}}$. Following the START~\citep{start}, we propose two hint instructions—\textit{Logical Verification} and \textit{Answer Reflection}—to insert tool-invoking hints into the original reasoning chains. As shown in Figure~\ref{fig:method1}, logical verification hints randomly replace uncertain expressions (e.g., \textit{maybe}, \textit{wait}, \textit{not sure}) in the chain, while reflection hints are inserted after the answer. These diverse hints facilitate the model to invoke tools when information is insufficient or after answer generation, enabling information completion and answer verification. After inserting hints, we truncate the original reasoning chain at the hint position, prompting the model to perform tool-augmented reasoning in response to the hint, which can be formulated as:
\begin{equation}
\small
P_{\theta}(\mathcal{R}^{\text{c}}_{>t^{H}}, y \mid I_{\mathcal{T}}, q, \mathcal{R}^{\text{c}}_{\leq t^{H}}, \mathcal{T}) = \underbrace{\prod\nolimits_{t=t^{H}}^{T_{c}} P_{\theta}(\mathcal{R}^{\text{c}}_t \mid \mathcal{R}^{\text{c}}_{\leq t}, I^{\mathcal{T}}, q, \{F^{\mathcal{T}}\}_{\leq t})}_{\text{Hint-based Tool-Integrated Reasoning}} \cdot \underbrace{\prod\nolimits_{t=1}^{T_y} P_{\theta}(y_t \mid y_{<t}, \mathcal{R}_{\text{c}}, I^{\mathcal{T}}, q)}_{\text{Answer Generation}},
\label{equ:reason_with_tools}
\end{equation}
Unlike Eq.(\ref{equ:reason_with_tools}), $t^H$ is the end position of the hint insertion, and $\mathcal{R}^{\text{c}}_{\leq t}$ represents the language-only reasoning chain before the hint position. The LLM continues reasoning from $\mathcal{R}^{\text{c}}_{\leq t}$ with $q$ to derive the tool-integrated reasoning chain $\mathcal{R}_{>t^{H}}^{\text{c}}$. Other variables are defined consistently with Eq.(\ref{equ:reason_with_tools}). After filtering the samples for the correct answers, we obtain the hint-based TIR dataset $D_{\text{tool}}^{H}$. Finally, we merge two datasets to obtain the final dataset $D_{\text{tool}}^{v1}= \{D_{\text{tool}}^{H} \cup D_{\text{tool}}^{M}\cup D_{\text{tool}}\}$.
\subsubsection{Step-2: Tool-Use Quality Normalization.} 
% Our aim is to construct a reasonable and efficient tool-use reasoning dataset. 
% Importantly, we observe significant noise in TIR trajectories, primarily due to tool overuse. 
To ensure the rationality of tool usage within each sample, we implement the following 3 TIR normalization strategies for tool-use data quality control: \textbf{(1) Tool-call Frequency Control:} Remove samples with tool-call frequency exceeding a predefined threshold $\beta$ to alleviate excessive tool invocation. \textbf{(2) Duplicate Tool-call Removal:} Eliminate samples containing redundant tool calls, such as repeated generation of identical search queries or code snippets in the same response. \textbf{(3) Format Normalization:} Standardize tool call formats in reasoning chains by unifying special tokens for invocation, feedback, and final answers, while ensuring balanced usage of start and end tokens. By applying these criteria, we obtain a quality-filtered dataset denoted as $D_{\text{tool}}^{\text{v2}}$.

\subsubsection{Step-3: Difficulty-Aware Data Classification.} 

Considering the computational overhead of tool-use and the multi-stage nature of TIR training, we argue that a high-quality tool-use dataset should meet the following criteria: \textbf{(1) Invoke tools only when necessary:} Tool calls should be avoided when the model is capable of solving the problem through direct reasoning. \textbf{(2) Organize samples from easy to hard:} As emphasized in prior RL tool learning~\citep{r1searcher,2504_DeepResearcher}, stage-wise training based on sample difficulty is crucial for effective learning.

To achieve the above objectives, we first perform a language-only reasoning pass on each question in the dataset $D_{\text{tool}}^{\text{v2}}$, yielding a direct reasoning result denoted as $D_\text{text}^{\text{v2}}$. Based on the correctness of both direct reasoning (DR) and tool-integrated reasoning (TIR), each sample is categorized into four types (Figure~\ref{fig:method1}). For categories 1 and 2, where the model already produces correct answers through direct reasoning, tool use is deemed unnecessary. These samples are extracted from $D_\text{text}^{\text{v2}}$ to form a subset $D_\text{text}^{\text{sub}}$. For category 3, which clearly illustrates the advantages of tool use over language-only reasoning, we directly sample from $D_{\text{tool}}^{\text{v2}}$ to construct the subset $D_{\text{tool}}^{\text{sub}}$.

To support a curriculum learning paradigm~\citep{how_abilities} from easy to hard, we construct a cold-start fine-tuning dataset $D_\text{tool}^{\text{SFT}}$ by combining $D_\text{text}^{\text{sub}}$ and $D_\text{tool}^{\text{sub}}$. For Category-4 samples, which are challenging for both DR and TIR, are treated as hard examples and reserved for reinforcement learning, forming the dataset $D_\text{tool}^{\text{RL}}$. This design enables the LLM to acquire basic tool-use capabilities through cold-start fine-tuning, and subsequently generalize to more complex scenarios during the RL phase (e.g. multi-tool collaboration), thereby facilitating a progressive learning trajectory.

\subsection{Tool-Star: A Two-stage Multi-tool Collaborative Training Framework}
\label{sec:tool_star_training}
In this section, we propose a two-stage multi-tool collaborative training framework that generalize the LLM's TIR capability from a single to multiple tool-use paradigm, and from easy to hard learning.

\textbf{Cold-Start Supervised Fine-tuning.} To equip the LLM with an initial understanding of tool usage for problem solving, given $(x_i, y_i) \in D_\text{tool}^{\text{SFT}}$, we apply the standard Supervised Fine-tuning objective on the backbone model $P_{\theta}$ with parameters $\theta$:
$
\label{eq:sft}
  \mathcal{L}(\boldsymbol{\theta})=-\sum_{(x_i, y_i) \in D_\text{tool}^{\text{SFT}}} \log P_{\theta}(y_i \mid x_i )
$
, where $x_i$ denotes the $i$-th input. Ultimately, we obtain a cold-start LLM $\hat \pi_{\theta}$ with initial TIR capability.

\textbf{Multi-Tool Self-Critic Reinforcement Learning.} We will introduce how the LLM learn to autonomously invoke tools, including a Code interpreter, a search engine and a web browser agent.

\begin{figure*}[!t]
\centering
\includegraphics[width=0.985\linewidth]{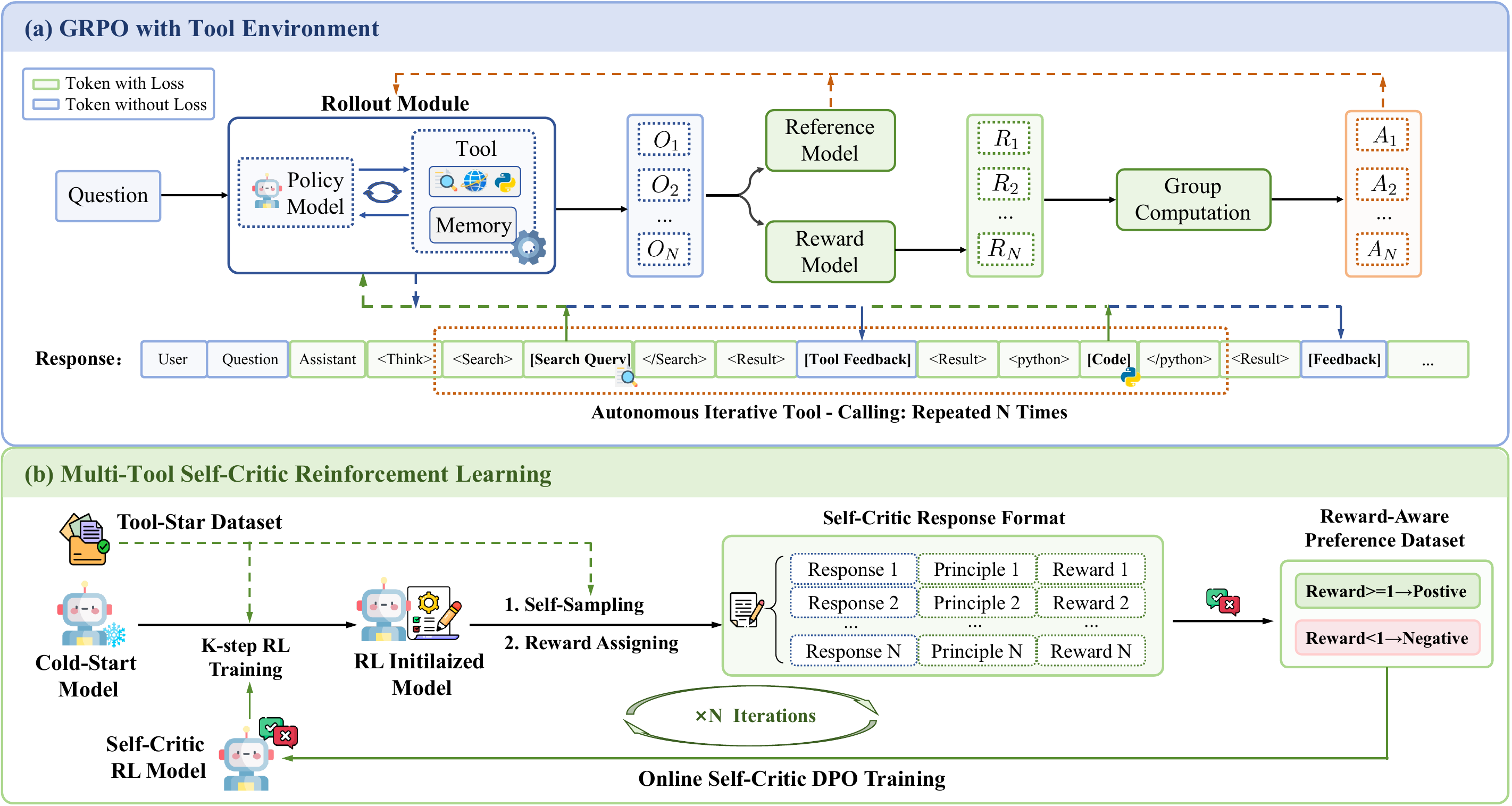}
\caption{
The overall framework of Multi-Tool Self-Critic Reinforcement Learning.
}
% \vspace{-1em}
\label{fig:method2} 
\end{figure*}

\textbf{(1) Memory-based Roll-Out with Tools.} As shown in Figure~\ref{fig:method2}, we employ multi-tool invocation instructions to guide the model in decoding tool-use requests into special tokens (e.g., \blue{<python>}) during the roll-out process. Upon detecting these tokens, the corresponding tool is automatically invoked, and the resulting feedback is integrated back into the reasoning chain. To reduce latency caused by frequent tool calls, we incorporate a memory mechanism that caches the mapping between each tool request and its output. This allows the model to retrieve responses for repeated requests directly from memory, thereby improving efficiency.

\textbf{(2) Hierarchical Reward Design.} Reward signals serve as the optimization objective and directly guide the behavior of the policy model during training. Distinct from previous tool-use RL approaches,  we not only design correctness and format rewards for LLMs, but also introduce multi-tool collaborative rewards. This design aims to encourage multiple tool usage while maintaining LLM correctness. Specifically, when both the answer and tool invocation format are correct, and the model employs multiple tools (i.e., both \green{<search>} and \blue{<python>} appear in the reasoning chain), an extra reward $r_{M}$ is granted. Formally, the overall reward $R$ is defined as follows:
\begin{equation}
\small
\hspace*{-0.5em} R = \begin{cases}
\max(Acc.+ r_{\text{M}}, Acc.) &\text{If Format is Good \& Acc.>0}  \\
0 & \text{If Format is Good \& Acc.=0 } \\
-1 & \text{Otherwise} \\
\end{cases}
,  r_{\text{M}} = 
\begin{cases}
0.1 & \text{If } \exists(\text{\green{<search>}} \&\text{\blue{<python>}}) \\
0 & \text{Otherwise}
\end{cases}
\end{equation}
\textbf{(3) Self-Critic RL Algorithm.} Multi-tool RL algorithms involve complex reward structures, making it challenging for LLMs to discover optimal behavior. To address this, we propose a self-critic RL algorithm that enables LLMs to better align with the design principles of reward mechanisms. As illustrated in Figure~\ref{fig:method2}, we begin by performing $K$ steps of vanilla RL training on the cold-start model $\hat \pi_{\theta}$. To optimize the policy, we adopt Group Relative Policy Optimization (GRPO)~\citep{deepseekmath} as our RL algorithm, which estimates the baseline using a group of rollouts.
\vspace{-0.2em}
\begin{equation}
\footnotesize
\begin{split}
    \hspace*{-0.6em} \mathcal{L}_{\text{GRPO}}(\theta) &= \mathbb{E}{[q \sim D_{\text{tool}}^{\text{RL}}, \{o_i\}_{i=1}^G \sim \hat \pi_{\theta_{old}}(O|q)]}  \frac{1}{G}\sum_{i=1}^G\frac{1}{|o_i|} \sum_{t=1}^{|o_i|}\\
    & \hspace*{-3.5em} \left\{ \min \left[ \frac{\hat \pi_\theta(o_{i,t} | q, o_{i,<t})}{\hat \pi_{\theta_{old}}(o_{i,t} | q, o_{i,<t})} \hat{A}_{i,t}, \text{clip} \left( \frac{\hat \pi_\theta(o_{i,t} | q, o_{i,<t})}{\hat \pi_{\theta_{old}}(o_{i,t} | q, o_{i,<t})}, 1 - \epsilon, 1 + \epsilon \right)  \hat{A}_{i,t} \right] - \beta \mathbb{D}_{KL}\left[\hat \pi_{\theta} || \hat \pi_{ref}\right]\right\} ,
\end{split}
\label{eq:grpo}
\end{equation}

where $\epsilon$ and $\beta$ are hyper-parameters, $q$ and $o$ denote query and model's output with tools. $A_{i}$ is the normalized advantage of the $i$-th rollout in group. Then we can obtain an RL-initialized model $\pi_\theta^{\text{RL}}$.

Subsequently, we detail our self-critic reward finetuning phrase, which help LLMs better internalize the reward structure. As shown in Figure~\ref{fig:method2}, we start by randomly rejection sampling $k$ examples from the RL training set $D_{\text{tool}}^{RL}$ to construct $D_{\text{tool}}^{\text{S}}$~\citep{llama,rft}. For each query $q \in D_{\text{tool}}^{\text{S}}$, the RL initialized model $\pi_\theta^{\text{RL}}$ self-samples $n$ candidate responses, forming a diverse QA set $D\sim(q, \{a_i\}_{i = 1}^{N})$. Notably, our hierarchical rule-based reward function acts as an executable program that automatically assigns a reward label to each response. This enables on-policy reasoning by providing each $a_i$ with a corresponding reward. We then construct a reward-aware dataset $D_{\text{tool}}^{\text{critic}} = \{(x_i, y_i)\}_{i = 1}^{N}$, where each input $x_i$ is a query $q$, and each output $y_i$ includes a candidate response $a_i$, its reasoning trace $p_i$, and reward score $r_i$. We treat samples with $r_i \geq 1$ as positive and those with $r_i < 1$ as negative, forming a preference dataset $(x, y_{w}, y_{l})\sim D_{\text{tool}}^{\text{critic}}$. Finally, we fine-tune the RL-initialized model using DPO objective as:
\begin{equation}
\mathcal{L}_{\text{DPO}}(\pi_\theta^{\text{RL}};\pi_\text{ref}) = - \mathbb{E}_{(x, y_{w}, y_{l})\sim \mathcal{D}}[\text{log}\sigma (\beta \text{log}\frac{\pi_\theta^{\text{RL}}(y_{w}|x)}{\pi_\text{ref}(y_{w}|x)}
- \beta \text{log}\frac{\pi_\theta^{\text{RL}}(y_{l}|x)}{\pi_\text{ref}(y_{l}|x)})],
\label{eq:dpo}
\end{equation}
% \end{gather*}
% \end{small}
% \end{equation}
where the reference model $\pi_\text{ref}$ is set to $\pi_\theta^{\text{RL}}$ initially and remains fixed throughout training. $\beta$ is a hyperparameter and $\sigma$ is the sigmoid function. $\mathcal{L}_{\text{DPO}}$ encourages higher log-probability for preferred responses $y_{w}$ over dispreferred ones $y_{l}$. We interleave self-critic reward fine-tuning every $k$ steps during standard RL training. This iterative process allows the model to progressively learn behavioral distinctions via self-sampling and evaluation, implicitly capturing the hierarchical reward structure. The full algorithm is detailed in Algorithm~\ref{algo:onpolicycriticrl}.

\subsection{Multi-tool Collaborative Inference}

During inference, we follow Eq.(\ref{equ:reason_with_tools}), enabling the model to autonomously use the \textit{search engine}, \textit{web browser agent}, and \textit{Python interpreter}. To improve TIR robustness, we introduce three inference-time mechanisms targeting common failure scenarios: 

\begin{itemize}[leftmargin=1em]

\item \textbf{Code Execution Error:} When generated code contains syntax errors, the \textit{Code Debugger} uses compiler messages and the original code to produce executable fixes. 

\item \textbf{Tool Invocation Failure:} If tool calls yield no useful output or debugging fails, the \textit{Backtracer} identifies the statement before the tool-invoking token and restarts inference from there. 

\item \textbf{Inference Length Overflow:} When tool outputs exceed the context limit, the \textit{Chain Refiner} compresses and restructures the reasoning chain by removing redundant or irrelevant content, allowing continued inference without tools. 
\end{itemize}
These mechanisms enhance tool invocation reliability and provide an effective inference-time scaling strategy. Further details are listed in the Appendix~\ref{app:tool_design}.

\section{Experiments}

\textbf{Datasets.} To comprehensively evaluate the tool-use capabilities of our model, we conduct experiments on two types of reasoning datasets:
(1) \textbf{Math reasoning benchmarks}, including AIME2024, AIME2025~\footnote{\url{https://huggingface.co/datasets/AI-MO/aimo-validation-aime}}, MATH500~\cite{math500}, MATH~\cite{MATH}, and GSM8K.
(2) \textbf{Knowledge reasoning benchmarks}, including three complex reasoning QA tasks: GAIA~\cite{GAIA}, HLE~\cite{HLE}, and WebWalker~\cite{2501_WebWalker}; and three Wikipedia-based open-domain QA tasks: HotpotQA~\cite{hotpotqa}, 2WikiMultihopQA~\cite{2wiki}, and Musique~\cite{musique}. For all tasks, we follow previous work~\cite{searcho1} and extract answers from the model output enclosed in \texttt{\textbackslash box\{\}}.

\textbf{Evaluation Metrics.} For computational reasoning tasks, we follow Search-o1~\citep{searcho1} employ LLM-based judging (Qwen2.5-72B=Instruct) to ensure answer correctness. For open-domain QA tasks, we adopt token-level F1 score as the evaluation metric. To assess tool usage efficiency, we propose the \textit{Tool-use Efficiency} metric: $T_E = \frac{1}{N} \sum_{i=1}^{N} \frac{S_i}{T^c_i}$ where $N$ is the number of datasets, $S_i$ and $T^c_i$ denote the number of correct answers and the number of total samples for the $i$-th dataset when using tools.

\textbf{Baselines.} Our experiments include the following three categories of baselines:
(1) \textbf{Proprietary Models}: Qwen2.5~\cite{qwen2.5}, Llama3.2~\cite{llama3}.  
(2) \textbf{Single-tool Enhanced Baselines}: Code-enhanced models such as ToRL~\cite{torl} and DotaMath~\cite{dotamath}, and search-enhanced models such as Vanilla RAG~\citep{rag_lewis}, Search-o1~\cite{searcho1} and Search-R1~\cite{searchr1}.  
(3) \textbf{Multi-tool Enhanced Baselines}: Multiple tools integrated prompting and ReCall~\cite{chen2025research}.
More details of implementations are provided in the Appendix~\ref{app:imp}.

\begin{table*}[!t]
\centering
\caption{Overall results on 10 challenging reasoning tasks. The top two results are highlighted in \textbf{bold} and \underline{underlined}. All baseline methods use Qwen2.5-3B-Instruct as the backbone. For baselines lacking a 3B Instruct version, we replicate them using the exact parameter settings from their open-source code. To ensure fairness, we report the RAG setting (Top-5 Docs) for the code-assistant model in knowledge-intensive reasoning. The abbreviations of the datasets: HQA (HotpotQA), 2Wiki. (2wikiMultiHopQA), MuSi. (MuSiQue), and Bamb (Bamboogle).  }
\label{tab:main_table}
\setlength\tabcolsep{1pt}
\renewcommand{\arraystretch}{1} % 修改行距
\fontsize{8.1pt}{10.5pt}\selectfont
\begin{tabular}{p{2.9cm}ccccccccccc}
\toprule
\multirow{2}[2]{*}{\textbf{Method}} & \multicolumn{5}{c}{\textbf{Computational Reasoning}} & \multicolumn{5}{c}{\textbf{Knowledge-Intensive Reasoning}} & \multirow{2}[2]{*}{\textbf{Avg.}} \\
\cmidrule(lr){2-6} \cmidrule(lr){7-11}
 & AIME24 & AIME25 & MATH500 & GSM8K & MATH & WebWalker & HQA & 2Wiki. & MuSiQ. & Bamb. & \\
\midrule
Qwen2.5-3B-Instruct & 10.0 & 6.7 & 63.0 & 75.0 & 71.6 & 0.5 & 9.7 & 9.4 & 3.6 & 11.7 & 26.1\\
Llama3.2-3B-Instruct & 0.0 & 3.3 & 40.0 & 71.2 & 58.2 & 0.5 & 12.5 & 9.2 & 4.0 & 18.3 & 21.7\\
\midrule
\multicolumn{12}{l}{\textit{\textbf{Code-Assistant Reasoning Methods}}} \\
ToRL  & \textbf{20.0} & \underline{10.0} & \textbf{72.0} & \underline{84.4} & \underline{81.0} & 12.0 & 37.9 & 27.0 & 8.3 & 25.4 & 37.8\\
DotaMath & 3.3 & 6.7 & 56.2 & 78.2 & 71.8 & 11.5 & 35.6 & 31.2 & 7.5 & 23.8 & 32.6\\
\multicolumn{12}{l}{\textit{\textbf{Search-Enhanced Reasoning Methods}}} \\
RAG & 13.3 & \underline{10.0} & 54.0 & 46.0 & 56.0 & 14.6 & 39.4 & 31.2 & 10.3 & 17.4 & 29.0\\
Search-o1 & \underline{16.7} & 3.3 & \underline{69.0} & 34.0 & 63.0 & 13.0 & 34.9 & 28.9 & 9.6 & 35.1 & 30.2\\
Search-R1 & 0.0 & 3.3 & 26.0 & 43.0 & 44.0 & 14.4 & 43.2 & 25.5 & 16.5 & 40.8 & 25.7\\
\midrule
\multicolumn{12}{l}{\textit{\textbf{Multi-Tool Integrated Reasoning Methods}}} \\
Multi-Tool Prompting & 3.3 & 3.3 & 54.2 & 48.8 & 59.6 & 9.2 & 15.9 & 16.2 & 6.5 & 17.8 & 23.4\\
ReCall & 16.6 &6.7 & 63.0 & 77.8 & 74.2 & 13.0& 43.5 & 38.9 & 16.5& 40.8& 39.1\\ 
\rowcolor[RGB]{236,244,252} 
Tool-Star(Qwen2.5-3B) & \textbf{20.0} & \textbf{16.7} & \textbf{72.0} & \textbf{85.0} & \textbf{82.6} & \underline{20.8} & \textbf{51.9} & \underline{40.0} & \textbf{19.3} & \underline{52.5} & \textbf{46.1} \\
\rowcolor[RGB]{236,244,252} 
Tool-Star(Llama3-3B) & 10.0 & \underline{10.0} & 54.8 & 77.8 & 70.6 & \textbf{24.0} & \underline{48.9} & \textbf{43.2} & \underline{16.6} & \textbf{54.7} & \underline{41.0}\\
\bottomrule
\vspace{-3em}
\end{tabular}
\end{table*}

\subsection{Results on Challenging Reasoning Tasks}

Our main results are presented in Table~\ref{tab:main_table}. Overall, Tool-Star consistently outperforms all baselines, decisively establishing the superiority. Furthermore, we have identified the following insights:
% \begin{enumerate}[leftmargin=1em]

\textbf{(1) TIR prompting fails to explore the better tool-use behavior.} Focusing on Search-o1 and Multi-Tool Prompting, their performance on comprehensive reasoning tasks remains suboptimal. Notably, multi-tool prompting strategies even exhibit lower consistency compared to their backbone model. This reveals that relying solely on prompt engineering to elicit tool usage is insufficient for guiding the LLM toward effective tool utilization, thereby limiting its reasoning capabilities.

\textbf{(2) Single-tool RL-based methods exhibit strong domain specialization but limited generalizability.} Search-enhanced RL approaches perform well on knowledge-intensive tasks (e.g., 40+\% F1 on HQA), but their accuracy drops sharply by over 20\% on computation-heavy tasks like MATH500 and AIME24 compared to the backbone model. Conversely, code-assistant methods like ToRL underperform on knowledge-based tasks, even falling behind RAG baselines. These trends underscore the specialization bias of single-tool RL methods and their limited cross-domain transferability.

\textbf{(3) Tool-Star demonstrates strong and versatile reasoning, along with robust plug-and-play capability.}
Tool-Star consistently outperforms both single- and multi-tool baselines, achieving over 40\% average accuracy across 10 datasets while maintaining competitive performance on individual tasks. Notably, it also delivers substantial gains across different backbone models, with average improvements nearing 20\% on both Qwen and LLaMA. These results highlight Tool-Star’s generality, efficiency, and strong adaptability across models and tasks. To further validate Tool-Star's generalizability, we present results on deep web exploration and a case study in Appendix~\ref{app:deep_web_explore} and~\ref{app:case}.

% \end{enumerate}

\subsection{Quantitative Analysis}

\begin{wraptable}{r}{0.52\textwidth}
\small
\vspace{-1.5em}
\centering
\renewcommand{\arraystretch}{1} % 修改行距
\caption{Ablation study on Tool-Star(3B).}
\vspace{-0.5em}
\label{tab:ablation}
\setlength\tabcolsep{1.7pt} % 调整列间距
\begin{tabular}{p{2cm}cccc}
\toprule
\multirow{2}[2]{*}{\textbf{Method}} & \multicolumn{2}{c}{\textbf{Knowledge.}} & \multicolumn{2}{c}{\textbf{Computational.}} \\
\cmidrule(lr){2-3}
\cmidrule(lr){4-5}
& HQA & Bamb. & GSM8K & MATH \\
\midrule
\rowcolor[RGB]{236,244,252} Tool-Star (3B) & 51.9 & 52.5 & 85.0 & 82.6 \\
\midrule
\textit{w/o} Cold-Start & 
$43.5_{\textcolor{deepred}{\,\text{(-8.4)}}}$ & 
$40.8_{\textcolor{deepred}{\,\text{(-11.7)}}}$ & 
$77.8_{\textcolor{deepred}{\,\text{(-7.2)}}}$ & 
$74.2_{\textcolor{deepred}{\,\text{(-8.4)}}}$ \\

\textit{w/o} RL stage & 
$47.5_{\textcolor{deepred}{\,\text{(-4.4)}}}$ & 
$43.9_{\textcolor{deepred}{\,\text{(-8.6)}}}$ & 
$80.2_{\textcolor{deepred}{\,\text{(-4.8)}}}$ & 
$78.4_{\textcolor{deepred}{\,\text{(-4.2)}}}$ \\

\textit{w/o} Reward & 
$50.4_{\textcolor{deepred}{\,\text{(-1.5)}}}$ & 
$50.3_{\textcolor{deepred}{\,\text{(-2.2)}}}$ & 
$83.1_{\textcolor{deepred}{\,\text{(-1.9)}}}$ & 
$80.2_{\textcolor{deepred}{\,\text{(-2.4)}}}$ \\

\textit{w/o} Self-Critic & 
$49.8_{\textcolor{deepred}{\,\text{(-2.1)}}}$ & 
$48.3_{\textcolor{deepred}{\,\text{(-4.2)}}}$ & 
$82.8_{\textcolor{deepred}{\,\text{(-2.2)}}}$ & 
$77.8_{\textcolor{deepred}{\,\text{(-4.8)}}}$ \\
\bottomrule
\end{tabular}
\vspace{-0.5em}
\end{wraptable}

\textbf{Ablation Study.} In this section, we conduct a systematic ablation study in Table~\ref{tab:ablation}, where “w/o” denotes removing a specific component. The results demonstrate that: \textbf{(1)} Removing any single module leads to performance degradation, highlighting the critical role of each part of design. \textbf{(2)} Excluding either results in significant performance drops.  the Cold-Start phase helps the model initially acquire tool-use capabilities, while the RL phase promotes generalization in multi-tool reasoning. Both stages are indispensable and complementary. \textbf{(3)} Incorporating hierarchical rewards and a self-critic mechanism on top of vanilla RL consistently brings further improvements, confirming the effectiveness of our tailored RL strategy.

% \end{figure*}
\begin{figure*}[!t]
\centering
\includegraphics[width=0.98\linewidth]{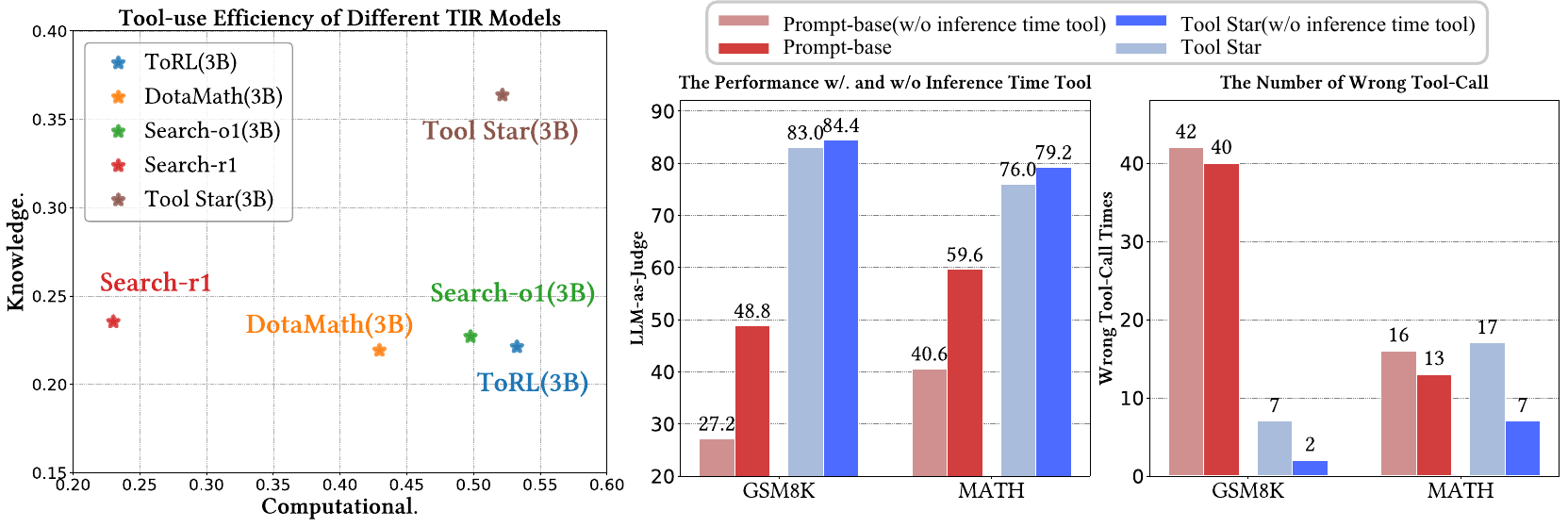}

\caption{
\textbf{Left:} Tool-use efficiency comparison across TIR methods. \textbf{Middle:} Performance analysis of TIR methods with inference-time tool. \textbf{Right:} Error statistics of tool invocation.
}
% \vspace{-0.5em}
\label{fig:tool_use_metric}
\end{figure*}

\textbf{Tool-Use Efficiency Analysis.} To validate whether Tool-Star can efficiently invoke tools during stepwise reasoning, we compare the \text{Tool-use Accuracy} $T_{E}$ of various TIR methods on knowledge-intensive and computation-based reasoning datasets, as shown in Figure~\ref{fig:tool_use_metric} (left). Results indicate that while Search-R1 and ToRL achieve high accuracy in knowledge and computation tasks respectively, they underperform in the other domain—highlighting the tendency of single-tool RL methods to specialize in specific reasoning types. In contrast, Tool-Star consistently maintains high tool-use accuracy across both task types and baselines, demonstrating its efficiency in tool utilization.

\begin{wrapfigure}[12]{r}{0.46\linewidth}
    \centering
    \vspace{-1.5em} % 上方间距
    \includegraphics[width=0.99\linewidth]{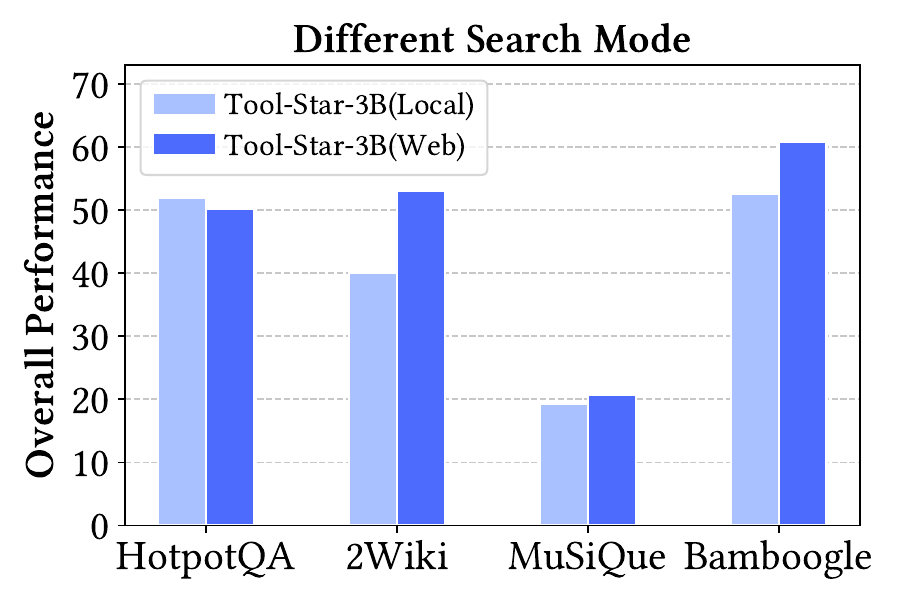}
    \vspace{-2em} % 下方间距
    \caption{Performance Comparison of Tool-Star(3B) in Local and Web Search Modes}

    \label{fig:search_compare}
\end{wrapfigure}

\textbf{Local Search VS. Web Search.} Inspired by R1-Searcher~\citep{r1searcher} and DeepResearch~\citep{2504_DeepResearcher}, we further compared the differences between local search and web search of Tool-Star-3B in knowledge-intensive reasoning tasks. As shown in Figure~\ref{fig:search_compare}, we found that after using web search, except for HotpotQA which did not show obvious improvement, the other three domains all demonstrated significant improvements, especially 2wiki (+13\% F1) and Bamboogle (+8\% F1). Regarding the above phenomena, we believe that the performance differences between the two mainly stem from: \textbf{(1)}The retrieval library of wiki is a subset of the web page corpus. Moreover, when combined with the browser agent for web page information, it can provide more accurate and concise information compared to wiki documents. \textbf{(2)} We use the Bing API~\footnote{\url{https://www.microsoft.com/en-us/bing/apis/bing-web-search-api}} for web search, and the retrieval models and strategies of their services perform better than the E5-base we deployed offline for local search.

\textbf{Inference-Time Tool-Use Analysis.} To verify the effectiveness of inference-time tool design, we further analyze its impact on the performance of Tool-Star and DotaMath across two datasets. As shown in Figures~\ref{fig:tool_use_metric} (middle) and (right), both models exhibit a notable reduction in tool-use errors during reasoning after applying inference-time tool optimizations, accompanied by significant performance improvements. These results not only underscore the detrimental effect of incorrect tool usage on reasoning accuracy, but also reinforces the motivation for incorporating inference-time tools. Notably, the performance gain for DotaMath is substantially larger than that for Tool-Star, with improvements exceeding 20\% on GSM8K. This suggests that inference-time tools are particularly beneficial for models with weaker coding capabilities, while models with stronger coding proficiency, such as Tool-Star, naturally make fewer tool-related errors and thus have less room for improvement.

\begin{wrapfigure}[12]{r}{0.48\linewidth}
    \centering
    \vspace{-1em} % 上方间距
    \includegraphics[width=1\linewidth]{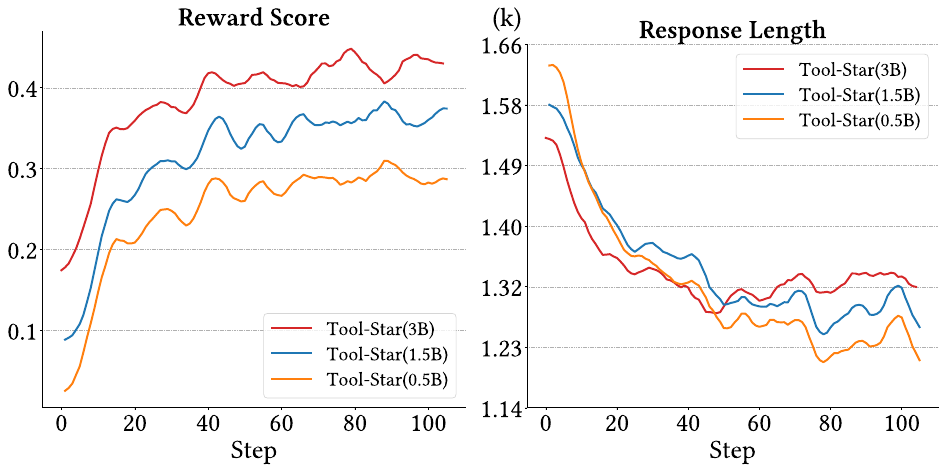}
    \vspace{-1.5em} % 下方间距
    \caption{Training curves of reward and response length for models with different parameters.}
    \label{fig:rl_training_curve}
    \vspace{-1.5em} % 下方间距
\end{wrapfigure}
\textbf{Scaling Analysis on Parameter Scales.} To investigate the effectiveness of Tool-Star across different parameter scales, we display the RL training curves of Qwen2.5 models with 0.5B, 1.5B, and 3B parameters in Figure~\ref{fig:rl_training_curve}. Our key observations are as follows: (1) All models achieve good reward scores at the beginning of training (step 0), benefiting from the cold-start SFT stage, which significantly reduces the exploration burden in early RL; (2) As training progresses, the average reward scores steadily improve across all model sizes, indicating that our self-critic RL framework further improve TIR capabilities on top of the cold-start initialization. (3) Notably, the average reward shows signs of emergent improvement around step 10, while no clear inflection point (i.e., “aha moment”) is observed in response length. Instead, response lengths gradually stabilize, suggesting convergence toward optimized response patterns. More training details can be found in Appendix~\ref{app:rl_mask}.

\begin{figure*}[!t]
\caption{The scaling relationship of kl loss, reward, response length and validation score for Tool-Star(Qwen2.5-3B \& 1.5B in instruct version).}
\centering
\includegraphics[width=1\linewidth]{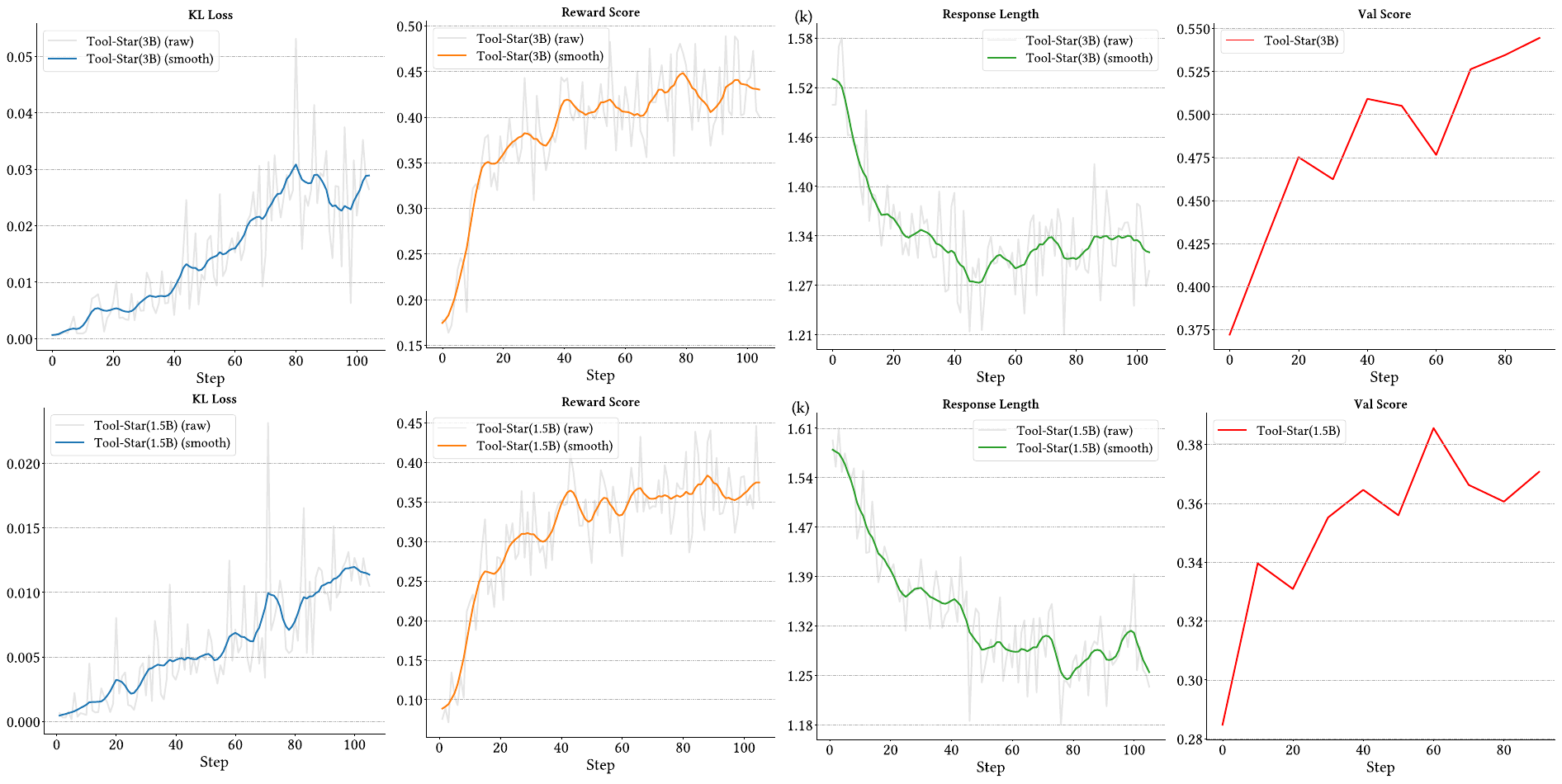}

\label{fig:train_step}
\end{figure*}

\textbf{Scaling Analysis on the RL Training Curve.}
\label{app:Analysis of the RL Training Curve} To better understand the training dynamics of the reinforcement learning process, we present the trends of KL penalty, average reward score, average response length, and validation score over the first 100 steps RL training (Figure~\ref{fig:train_step}). Our key observations are as follows:
\begin{itemize}[leftmargin=1em]
\item All metrics—including reward score, KL penalty, and validation score—increase steadily with training steps, indicating the stability of our self-critic RL training method.

\item The initial reward scores of the 3B and 1.5B models are already above 0 (within a possible range of -1 to 1.1), with corresponding validation scores starting at approximately 0.375 and 0.29 respectively—substantially higher than the vanilla instruct model's score of below 0.15. These results demonstrate the effectiveness of our cold-start strategy. In contrast to RL methods trained entirely from scratch~\citep{torl,r1searcher,searchr1}, the cold-start approach endows the model with a baseline level of TIR capability from the outset, significantly reducing the need for extensive black-box exploration during the early stages of RL training.

\item After cold-start SFT, the reward score shows mild emergence within the first 0–20 steps, though less pronounced than in from-scratch RL (i.e., no sudden rise-and-drop). Similarly, the response length does not fluctuate sharply but gradually converges to a stable value, further supporting the stability benefits of cold-start training.
\end{itemize}

\section{Conclusion}

In this paper, we introduce Tool-Star, an RL-based framework designed to enable LLMs to autonomously invoke multiple external tools during stepwise reasoning. To address the scarcity of tool-use data, Tool-Star employs a scalable TIR data synthesis pipeline, incorporating normalization and a difficulty-aware data classification process. Furthermore, we propose a two-stage training framework to enhance multi-tool collaborative reasoning in LLMs, consisting of a Cold-Start Fine-tuning phase and a Multi-Tool Self-Critic RL stage. This framework progressively fosters effective multi-tool collaboration and strengthens the LLM’s understanding of reward principles. Extensive experiments across over ten challenging benchmarks demonstrate that Tool-Star significantly improves both the effectiveness and efficiency of tool-augmented reasoning in LLMs.

{
\bibliographystyle{plain} % plain, unsrt
\bibliography{main}
}

\clearpage
\appendix

\section*{Appendix}
\startcontents[sections]
\printcontents[sections]{l}{1}{\setcounter{tocdepth}{3}}

\clearpage

\section{Supplement Experimental Results}

\begin{table*}[t]
\centering
\caption{The performance of Tool-Star and other baseline methods on the GAIA and HLE datasets. The experimental settings are consistent with those in Table \ref{tab:main_table}. Here, Qwen refers to the use of the Qwen2.5-3B-Instruct model, and Llama refers to the use of the Llama3.2-3B-Instruct model.}
\label{tab:gaia_hle}
\setlength\tabcolsep{1pt}
\renewcommand{\arraystretch}{1} % 修改行距
\fontsize{9.5pt}{11.5pt}\selectfont
\begin{tabular}{p{1.5cm}ccccccccc}
\toprule
\multirow{1}[0]{*}{\textbf{Dataset}} & Qwen & Llama & ToRL & DotaMath & RAG & Search-o1 & Search-R1 & Multi-Tool Prompting & Tool-Star \\
\midrule
GAIA & 3.9 & 5.8 & 8.7 & 4.9 & 8.7 & 7.8 & \underline{9.7} & 5.8 & \textbf{15.5}\\
HLE & 7.4 & 5.0 & 7.2 & 4.8 & 7.2 & 6.2 & \underline{7.8} & 5.6 & \textbf{8.0}\\
\bottomrule
\vspace{-3em}
\end{tabular}
\end{table*}

\subsection{Deep Web Exploration Result}
\label{app:deep_web_explore}
In this section, we further evaluate the performance of the Tool-Star framework on web exploraton tasks. Specifically, we assess it on the \textbf{General AI Assistants (GAIA)} and \textbf{Humanity's Last Exam (HLE)} datasets with the same experimental settings as in main result. 

As shown in Table~\ref{tab:gaia_hle}, In detail, we observe that the Tool-Star framework achieves strong performance even on web search tasks. Specifically, on the GAIA dataset, our method achieves the highest score, outperforming the second-best by over 60\%. On the HLE dataset, our method also demonstrates strong competitiveness. This fully validates the effectiveness of our approach in the deep web exploration task. 

In terms of task difficulty, GAIA consists of complex deep web exploration tasks that typically require both information retrieval and reasoning. To address such challenges, Tool-Star often employs a multi-tool collaboration strategy to accomplish the subtasks. Examples of multi-tool collaborative reasoning are shown in Table~\ref{tab:case_gaia}.

\begin{wraptable}{r}{0.48\textwidth}
\centering
\vspace{-1em}
\caption{Performance comparison of RL Algorithm on ToolStar (Qwen2.5-3B-Instruct).}
\label{tab:toolstar_rl}
\vspace{-0.5em}
\setlength\tabcolsep{2pt}
\renewcommand{\arraystretch}{1.2}
\fontsize{9pt}{11pt}\selectfont
\begin{tabular}{lcccc}
\toprule
\textbf{Models} & GSM8K & MATH &HQA. & Bamb. \\
\midrule
Tool-Star (SFT)& 80&79.2 &47.5 &48.3 \\
\hline
+ GRPO & 85.0 & 81 & 51.9 & 52.5 \\
+ REINFORCE++  & 84.4 & 80.5 & 48.8 & 53.1 \\
\bottomrule
\end{tabular}
\vspace{-1em}
\end{wraptable}

\subsection{Different RL Algorithm: REINFORCE++ VS. GRPO}

As two widely-used rule-based reinforcement learning algorithms, Tool-Star with GRPO~\citep{deepseekmath} and Reinforce++~\citep{hu2025reinforce++} are compared in Table~\ref{tab:toolstar_rl} in terms of training curves and experimental performance. It can be seen that the overall differences between GRPO and Reinforce++ are relatively minor.

Experimental results show that Tool-Star with GRPO slightly outperforms REINFORCE++ on computational reasoning tasks such as GSM8K and MATH. However, when it comes to knowledge-intensive reasoning tasks, REINFORCE++ slightly outperforms GRPO on Bamboogle. These results suggest that while the performance of the two methods fluctuates slightly across different domains, the overall difference remains within 3\%.

\subsection{Analysis of Tool-Call Masking}
\label{app:rl_mask}
\begin{wrapfigure}[12]{r}{0.4\linewidth}
    \centering
   \vspace{-2em} % 上方间距
    \caption{The result of tool-call masking}
     \vspace{-0.5em} % 上方间距
    \includegraphics[width=0.8
    \linewidth]{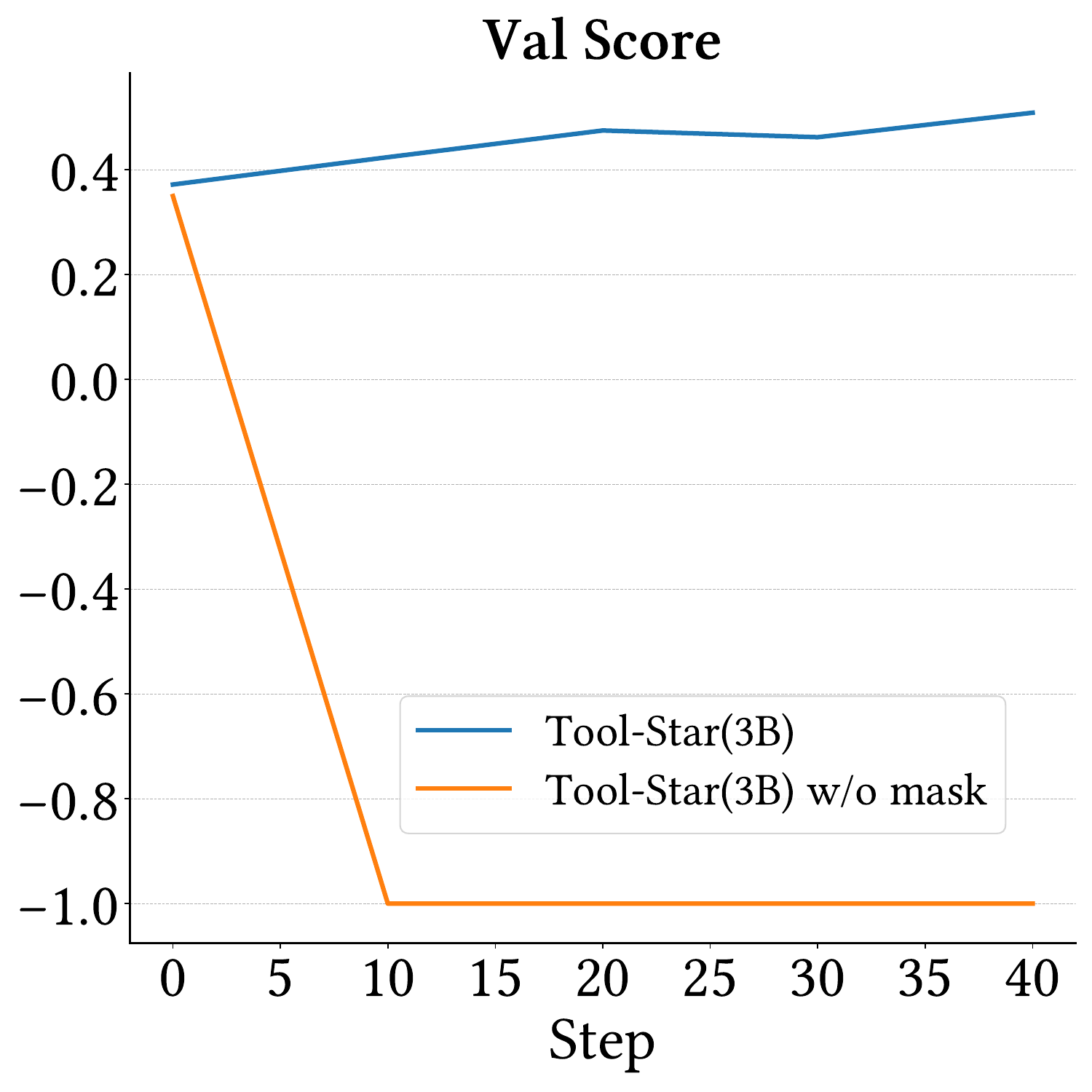}
    \label{fig:rl_mask}
    % \vspace{1em} % 下方间距
\end{wrapfigure}
In our work,  we apply masking to the retrieved and python execution results when computing the loss to prevent the training policy from being biased by tool outputs. Therefore, all tool call results in Tool-Star are masked. Based on this design, we conduct ablation studies on Tool-Star (Qwen2.5-3B-Instruct) with or without tool-call masking in RL training.

As shown in Figure~\ref{fig:rl_mask}, removing the masking leads to a rapid collapse of the model's reward score to -1, with persistent strategy hacking. This stands in contrast to prior findings in single-tool reinforcement learning~\citep{zhang2025agent}, where, according to their experiments, masking in single-tool RL did not lead to result hacking. We consider that maybe tool feedback in multi-tool reasoning can introduce significantly greater instability during RL training.

\section{Details of Tool-Star Framework}

\begin{algorithm}[t]
\caption{Multi-Tool Self-Critic Reinforcement Learning}\label{algo:onpolicycriticrl}
\begin{algorithmic}[1]
\Require Reasoning Model $\mathcal{\pi_\theta}$, External Tools $T$, Reward Function $\mathcal{R_\phi}$, Cycle Times $\mathcal{C}$, GRPO Steps $\mathcal{S}$
% , Clipping Parameter $\mathcal{\epsilon}$, KL Penalty Coefficient $\mathcal{\beta}$, Learning Rate $\mathcal{\eta}$
\State \textbf{Input:} Datasets $D$, Task instruction $I$
\State Initialize ${\pi_\theta}_\text{old} \gets \pi_\theta$
\For{$i \gets \{1, ..., \mathcal{C}\}$}
    \For{$\text{step} \gets \{1, ..., \mathcal{S}\}$}{\color{blue}\Comment{Start Vanilla RL phrase (GRPO)}}
        \State Sample a batch $D_b$ from $D$
        \For{each query $q \in D_b$}
            \State $q \gets I\oplus q$
            \State Sample $G$ sequences for $q$ with $T$: $\{o_j\}_{j=1}^G \sim {\pi_\theta}(\cdot \mid q, T)$
            
            \State Calculate the optimization objective function $J_\theta$ of GRPO according to Eq.(\ref{eq:grpo})
            \State Update the weight parameters of $\mathcal{\pi_\theta}$: $\theta \gets \theta + \eta 
 \cdot \nabla_\theta J_\theta$
           
        \EndFor
    \EndFor 
    \State Randomly sample $D_{sample}$ from $D$ {\color{blue}\Comment{Start Self-Critic DPO}}
    \For{each query $q \in D_{sample}$}
        \State $q \gets I\oplus q$
        \State Sample the query $q$ for $G$ times with $T$ to generate $G$ different results: $\{o_j\}_{j=1}^G \sim {\pi_\theta}(\cdot \mid q, T)$
        \State Filter one correct result and one incorrect result from $G$ results to obtain $o_{chosen}$ and $o_{reject}$
        \State Calculate the optimization objective function $J_\theta$ of DPO according to Eq.(\ref{eq:dpo})
        \State Update the weight parameters of $\mathcal{\pi_\theta}$: $\theta \gets \theta + \eta 
 \cdot \nabla_\theta J_\theta$
    \EndFor
\EndFor
\State \textbf{Output:} Trained Model $\mathcal{\pi_\theta}_\text{new}$
\end{algorithmic}
\end{algorithm}

\subsection{Training Algorithm of Tool-Star}

In this section, we delve into the overall workflow of the our Multi-Tool Self-Critic Reinforcement Learning algorithm. The training process consists of two alternating phases over multiple cycles, which are illustrated in algorithm~\ref{algo:onpolicycriticrl}:

\textbf{(1) Vanilla RL stage:} In each cycle, the model $\pi_\theta$ is first optimized via GRPO for a fixed number of steps. For each query sampled from the dataset, the task instruction is prepended, and the model generates multiple output sequences using external tools. These outputs are used to compute the GRPO objective (Eq.~\ref{eq:grpo}), and the model parameters are updated accordingly.

\textbf{(2) Self-Critic DPO stage:} After GRPO, a self-critic phase is performed to refine the model's decision-making. For each sampled query, the model generates multiple outputs, from which one preferred (correct) and one rejected (incorrect) response are selected. These are used to compute the DPO objective (Eq.~\ref{eq:dpo}), further updating the model via gradient ascent.

This alternating training scheme enables the model to not only improve through reinforcement learning with tool-augmented outputs but also self-correct by distinguishing between good and bad generations. The final output is a tool-aware reasoning model $\pi_\theta^\text{new}$ optimized for both performance and efficiency.

\subsection{Details of Tool Design}
\label{app:tool_design}
To provide a more detailed illustration of the design and usage guideline of our tools, we present the specifics of both the training tools and inference-time tools below.

\textbf{Training Tools:}
\begin{itemize}[leftmargin=1em]
\item \textbf{Search Engine:} During inference, the search engine executes queries to retrieve relevant information, supporting both local and web-based retrieval modes. Inspired by DeepResearch~\citep{2504_DeepResearcher}, we adopt web search during training. Empirically, we find that both using webpage snippets and incorporating a browser agent consistently improve RL performance. However, the former significantly reduces efficiency. During evaluation, we use Wikipedia\footnote{https://archive.org/details/enwiki-20181220} as the retrieval corpus and E5~\citep{wang2024textembeddingsweaklysupervisedcontrastive} as the retriever for four open-domain QA tasks. For all other tasks, we combine web search with the browser agent to obtain higher-quality retrieval results.

\item \textbf{Web Browser Agent:} This component accesses URLs from search results, parses webpage content, extracts information relevant to the query, and summarizes it. The process includes webpage parsing, regex-based cleaning, and an auxiliary model that refines the cleaned content based on the search query. The summarized output is then fed back into the reasoning chain. Notably, the browser agent is optional—when it is not used, the search engine directly returns webpage snippets or raw text.

\item \textbf{Code Interpreter:} Code snippets generated by the language model are executed in a sandboxed environment, returning either the execution result or error messages based on correctness. Our Python interpreter implementation follows the design of ToRA~\citep{tora}, ensuring secure and accurate execution.
\end{itemize}
\textbf{Inference-time tool:} 

\begin{itemize}[leftmargin=1em]

\item \textbf{Code Debugger:} This module automatically corrects erroneous code generated by large language models and guides the code revision process. Specifically, we design a dedicated debug prompt (~\ref{app:instruction_debugger}) that incorporates the model-generated code and compiler error messages, enabling the model to understand and fix the errors for automated code correction.

\item \textbf{Tool-Use Backtracer:} This module identifies and reverts to the reasoning step prior to a failed tool call, allowing the model to recover and revise its reasoning path. In real experiments, we compared model-based and rule-based rewinding strategies. As the model often struggles to accurately locate the faulty tool call, we adopt the rule-based approach. Concretely, we set the rewind position at the newline token preceding the special token of the erroneous tool call, prompting the model to regenerate the reasoning chain from that point.

\item \textbf{Reasoning Chain Refiner:}
When the model output exceeds the maximum length limit, this module prunes and refines redundant steps in the reasoning process, replacing the original chain with a more concise and coherent version to complete the task. We design a refinement prompt (~\ref{app:instruction_refiner}) that enables the model to identify and remove errors or repetitive thoughts in the original reasoning chain and generate a streamlined version. Template examples can be found in Table~\ref{app:instruction_refiner}.
\end{itemize}
To ensure fairness, all deployed models of tools in the experiments (e.g., the browser agent's information refinement process, code debugger, and Reasoning Chain Refiner) use instruct-version models with the same parameter size as the test model. Therefore, our inference-time tool can also be regarded as a self-play inference-time scaling process.

\begin{table}[t] % r 表示靠右，0.6 表示宽度
    \caption{The statistics of data sources in our main result.}
    \label{tab:data_source}
    \centering
    \small  
    \setlength{\tabcolsep}{10pt} % 可微调列间距
    \begin{tabular}{lccc}
        \toprule
        \textbf{Dataset} & \textbf{Type} &\textbf{Task} & \textbf{\# Counts} \\
        \midrule
        % AoPS Forum & Computational Reasoning & 30.2k \\
        % AMC\&AIME(Before 2024) & Computational Reasoning & 4.6k \\
        % cn\_k12 & Computational Reasoning & 276.6k \\
        % GSM8K~\citep{cobbe2021gsm8k} & Computational Reasoning & 7.3k \\
        % MATH~\citep{bigmath} & Computational Reasoning & 12.0k \\
        % Olympiads & Computational Reasoning & 150.6k \\
        % Orca-Math & Computational Reasoning & 153.3k \\
        % Synthetic-AMC & Computational Reasoning & 62.1k \\
        NuminaMath & Language-only& Computational Reasoning &  72.4K \\
        HotpotQA~\citep{r1searcher} & Language-only& Knowledge-Intensive Reasoning & 8.1k \\
        2WikiMultiHopQA~\citep{2wiki} & Language-only& Knowledge-Intensive Reasoning & 10k \\
        SimpleDeepSearch~\citep{SimpleDeepSearcher} & TIR & Knowledge-Intensive Reasoning & 0.9k \\
        WebWalker~\citep{2501_WebWalker} & Language-only& Knowledge-Intensive Reasoning & 3k \\
        \bottomrule
    \end{tabular}
\end{table}

\section{Implementation Details}
\label{app:imp}

\subsection{The Statistics of Data Sources}
\label{app:data_collection}
In this section, we sample data from sources including Numina-MATH, 2Wiki, HotpotQA, with the detailed breakdown and corresponding data sizes presented in Table~\ref{tab:data_source}.
We strictly adhere to all copyright and licensing terms associated with the datasets and open-source code.

For the computational reasoning dataset, we use 72.4K samples aligned with Numina-MATH TIR. For knowledge-based reasoning, we include 10.6K samples from HotpotQA and 2Wiki, aligned with R1 Searcher and Research respectively, along with 0.9K samples from the SimpleDeepSearch dataset and 3K samples randomly selected from the Webwalker training set. Finally, we obtained approximately 100K samples containing 1\% TIR data.

It is important to note that we \textbf{only use the queries} from these language-only datasets for reasoning chain sampling, without relying on the reasoning chains generated by stronger models of original datasets. Our cold-start data is sampled using Qwen2.5-3B-Instruct, which serves as the backbone model to match the parameter scale of the main table. Following the START settings~\citep{start}, we set the temperature to 0.7 and the top-p value to 0.95 during sampling. Due to resource constraints, each question is sampled three times only. In the tool-use quality normalization stage, our tool-call frequency threshold $\beta$ is set to 5. 

Ultimately, following multiple stages of quality filtering in Section~\ref{sec:tool_data_synthesis}, we curated approximately 54K samples from categories 1, 2, and 3 for cold-start training. Furthermore, 10K challenging samples were randomly selected from categories-4 samples for the RL phase to enable progressive alignment from simpler to more difficult tasks.

\subsection{Training Details of Tool-Star}
\paragraph{Cold-Start Supervised Finetuning.}  
After applying our sampling strategies to the original datasets, we use approximately 54k samples for cold-start supervised fine-tuning. During the supervised fine-tuning stage, we fine-tune a Qwen2.5-3B-Instruct model on the training data with a learning rate of 7e-6. We utilize DeepSpeed ZeRO-3~\citep{deepspeed} and FlashAttention2~\citep{flashattention} for training. The batch size is set to 128, weight decay to 0.1, and we train for 3 epoch. Mixed precision training with BF16 is adopted, and the maximum input length is set to 4096.

\paragraph{Multi-Tool Self-Critic Reinforcement Learning.}  
After SFT, we further train the model using the GRPO algorithm under the VERL framework~\footnote{\url{https://github.com/volcengine/verl}}. Similarly, we randomly select around 10k samples from $D_{tool}^{RL}$ for reinforcement learning. The learning rate is set to 8, with 8 rollouts per sample. The total training batch size is 128, and the mini-batch size is 16. The maximum output length is set to 4096 to ensure completeness of exploration. To accelerate the training speed, following DAPO~\citep{DAPO}, we set the maximum number of tool invocations during training to 3. To stabilize training, we set the KL divergence coefficient in GRPO to 0. The reinforcement learning phase lasts for 2 epochs. All experiments are conducted on 8 NVIDIA A800 GPUs. To ensure reproducibility, all experimental results are averages of three independent trials. We have approximately 300 samples in the validation set during the RL stage, which includes AIME25, 25, MATH500, HotpotQA, 2wikiMultiHopQA, MuSiQue and Bamboogle.  

It is worth noting that \textbf{all tool invocation results in Tool-Star are masked}. As illustrated in the Figure~\ref{fig:method2}, these masked segments are excluded from the loss computation to prevent the training policy from being biased toward the tool outputs. Only the tokens from the text-based reasoning process and tool requests are considered during loss calculation.

% 我们在RL阶段的validation set约300条，包含AIME25，25，MATH500,HotpotQA,2wikiMultiHopQA),MuSiQue and Bamboogle.

\paragraph{Self-Critic DPO Training.} In this phase, the learning rate is set to 5e-7 with a cosine scheduler and a 0.1 warm-up ratio. We use DeepSpeed ZeRO Stage 3 and Flash-Attention 2 for efficiency, with a global batch size of 64. Training utilizes a sigmoid loss function with a beta value of 0.3 and spans 2 epochs. Mixed precision training with bf16 is employed, and the maximum context length is 4096 tokens. We perform self-critic training every 0.5 epoch during the standard reinforcement learning process (approximately every 50 steps), and use the resulting model to resume the subsequent vanilla reinforcement learning training. The detailed data format are listed in Appendix~\ref{app:Cases of Self-Critic DPO Data}

\subsection{Details of Retrieval}
During training, we use the Bing Web Search API as the retriever, configured with the US-EN region setting. For each query, we retrieve 10 web pages as supporting documents.
During inference, we use Bing Web Search for the Math Reasoning and Complex Reasoning datasets. For open-domain QA, we employ Wikipedia as the retriever. The Wikipedia retriever is based on the Wikipedia 2018 Corpus provided by FlashRAG~\cite{flashrag} and uses E5-base-v2 as the retriever.

\subsection{Baseline Details}

\textbf{Proprietary Models}
\begin{itemize}[leftmargin=1em]

\item \textbf{Qwen2.5 Series}~\citep{qwen2,qwen2.5}, developed by Alibaba, is available in several sizes: Qwen2.5-0.5B /1.5B /7B and 72B. This model is trained on data sources spanning 29 kinds of languages, enabling it to perform exceptionally well in multilingual tasks. Additionally, Qwen2.5 exhibits strong capabilities in coding and mathematics. Qwen2.5-72B-Instruct is notable for its ability to handle input windows of up to 128K tokens in length, making it exceptionally well-suited for processing long texts and tackling complex tasks.

\item \textbf{LLaMA 3 Series}~\citep{llama3}, developed by Meta AI, is released in multiple sub-series, including LLaMA 3, LLaMA 3.1, LLaMA 3.2 and LLaMA 3.3. LLaMA series models offer two primary variants for each model size: the base model and the instruct model. These models are trained on a mixture of datasets, covering a wide range of domains and languages. The LLaMA series models achieve strong performance across reasoning, coding, and multilingual understanding benchmarks. With continued improvements in alignment and context length, they are increasingly suitable for real-world tasks.
\end{itemize}

\textbf{Search-Enhanced Baselines}
\begin{itemize}[leftmargin=1em]
\item \textbf{RAG~\citep{rag_lewis}} is a basic baseline that demonstrates the model's ability to directly utilize retrieved information. In our implementation, we directly used the retriever to obtain the top-5 results and inserted them into a carefully designed prompt to let the model generate the answer.
    
\item \textbf{Search-o1~\citep{searcho1}} is a reasoning-model-driven RAG method. Its core idea is to allow the reasoning model to autonomously decide whether to perform a search during its reasoning process. It generates queries, retrieves relevant information, organizes the retrieved results, and incorporates them into the model’s reasoning chain. The model then continues generation based on this chain until a final answer is reached. In our experiments, we modified the official code of Search-o1 by replacing its base model with one that has the same number of parameters as ours, and we also used the same retriever to ensure consistency.

\item \textbf{Search-R1~\citep{searchr1}} is a reinforcement learning method that trains a model from scratch to call the search engine during reasoning. Its overall process is similar to Search-o1~\citep{searcho1}: the model generates necessary queries during reasoning, retrieves relevant information, and inserts it into the reasoning chain. We follow the original prompt used in the method to ensure its effectiveness.
\end{itemize}

\textbf{Code-Enhanced Baselines}
\begin{itemize}[leftmargin=1em]
\item \textbf{ToRL~\citep{torl}} is a tool-integrated framework trained with reinforcement learning methods. This framework enables the model to autonomously invoke code interpreter to solve mathematical reasoning problems. Compared to prompt-based methods and supervised fine-tuning training approaches, ToRL encourages the model to explore extensively to find optimal tool usage strategies. This approach shows significant improvements over traditional methods. At the same time, the framework demonstrates the broad potential applications of tool integration in mathematical reasoning.
    
\item \textbf{DotaMath~\citep{dotamath}} models are fine-tuned on mathematical instruction datasets. After training, the DotaMath model demonstrates strong capabilities in solving complex mathematical reasoning problems. By training on the DotaMathQA dataset built for mathematical problems, the model has acquired the abilities of thought decomposition and self-correction during reasoning. This paradigm opens up a new direction for solving complex math problems.
\end{itemize}

\textbf{Multi-tool Enhanced Baselines}
\begin{itemize}[leftmargin=1em]
\item \textbf{Multi-Tool Prompting} method refers to the approach where the model learns to use external tools to solve problems solely through prompts. In this setup, we primarily aim to explore whether the model has the capability to autonomously call relevant tools to complete tasks. For the prompt format, we use the same structure as the one employed by Tool-Star. The specific prompt can be found in Appendix~\ref{app:instruction_tool_star}.
    
\item \textbf{ReCall~\citep{chen2025research}} method makes significant improvements over the ReSearch framework. Specifically, compared to the purely search-based framework, it further allows the model to call code interpreter to solve complex reasoning problems. This setup expands the tasks that ReSearch could handle to mathematical reasoning, while still maintaining the original capability of ReSearch to call search engines. Additionally, ReCall introduces a new method for generating data towards complex multi-step tasks, enabling the LLM to develop sophisticated reasoning abilities based on multiple tools.
\end{itemize}

\subsection{Datasets Details}
We provide a detailed introduction to the computational reasoning benchmarks evaluated by Tool-Star:

\begin{itemize}[leftmargin=1em]
\item \textbf{AIME24~\footnote{\url{https://huggingface.co/datasets/HuggingFaceH4/aime_2024}}} A dataset of 30 challenging math problems from the 2024 American Invitational Mathematics Examination (AIME I \& II), covering algebra, geometry, number theory, and combinatorics. Each JSONL-formatted entry includes the problem ID, statement, detailed solution steps, and the final integer answer. Designed to evaluate large language models on complex mathematical reasoning, the dataset features high difficulty, rigorous multi-step solutions, and unique answers—making it a strong benchmark for assessing advanced problem-solving capabilities.

\item \textbf{AIME25~\footnote{\url{https://huggingface.co/datasets/math-ai/aime25}}} is a dataset comprising 30 challenging mathematical problems sourced from the 2025 American Invitational Mathematics Examination (AIME I \& II). These problems encompass a diverse range of topics, including algebra, combinatorics, geometry, number theory, and probability. AIME25 is designed to evaluate the mathematical reasoning capabilities of LLMs. It serves as a benchmark for assessing how well these models can tackle complex mathematical problems requiring multi-step reasoning and understanding of advanced mathematical concepts.

\item \textbf{GSM8K~\citep{cobbe2021gsm8k}} is a widely adopted benchmark for assessing the mathematical reasoning abilities of large language models through grade-school level word problems. In our experiments, we utilize its standard test set and report performance based on greedy decoding accuracy (maj@1).

\item \textbf{MATH~\citep{MATH}} is a dataset consisting of challenging high school-level mathematics problems, spanning topics such as Prealgebra, Algebra, Number Theory, Counting \& Probability, Geometry, Intermediate Algebra, and Precalculus. Compared to GSM8K, MATH offers greater difficulty and topic diversity. 

\item \textbf{MATH500~\citep{math500}} is a subset of 500 problems drawn from the MATH benchmark developed by OpenAI. This dataset is designed to assess the mathematical reasoning capabilities of large language models. The problems in MATH500 cover a range of mathematical topics including algebra, geometry, trigonometry, and probability. Each entry in the dataset includes the problem statement, a detailed step-by-step solution, and the final answer, allowing for thorough model training and evaluation.

\end{itemize}

Furthermore, we also conduct a series experiments on follow knowledge-intensive reasoning benchmarks:

\begin{itemize}[leftmargin=1em]
\item \textbf{WebWalker~\citep{2501_WebWalker}} s a dataset consisting of queries derived from various web pages, where answering a single query may require gathering information from one or multiple pages. In our experiments, we adopt an open-domain setting by using a search engine for retrieval rather than the root URLs provided in the dataset.
\item \textbf{HotPotQA~\citep{hotpotqa}} is a widely-used question answering dataset that features multi-hop questions built from Wikipedia articles. In our experiments, we use its validation set for evaluation.
\item \textbf{2WikiMultihopQA~\citep{2wiki}} is a large-scale multi-hop QA dataset that combines structured knowledge from Wikidata and unstructured text from Wikipedia. It includes diverse question types and annotated reasoning paths to support explainable multi-step inference.
\item \textbf{Bamboogle~\citep{bamboogle}} is a challenging multi-hop QA dataset that contains hand-crafted questions requiring up to four reasoning hops. It tests a model's ability to perform complex and compositional reasoning across multiple facts.
\item \textbf{Musique~\citep{musique}} is a multi hop dataset that requires inference. The problem inside is a dependency based multi hop problem, where one inference step depends on the information obtained from the previous inference step, making the dataset very difficult. We conducted experiments using Musique's dev set.
\end{itemize}

\section{Instruction Templates}
\label{app:instructions}

\subsection{Instructions for Tool-Star}
\label{app:instruction_tool_star}

\begin{tcolorbox}[
    colframe = gray,       % 边框颜色
    colback = gray!5!white,             % 背景颜色
    coltitle = white,                   % 标题字体颜色
    coltext = black,                    % 文字颜色
    fonttitle = \bfseries,              % 标题字体加粗
    title = Instruction for Tool-Star,  % 标题内容
    boxrule = 1pt,                      % 边框宽度
    arc = 2mm,                          % 边角圆润
    width = \linewidth,                 % 宽度
    left = 7pt,                         % 左边距
    right = 7pt,                        % 右边距
    top = 5pt,                          % 上边距
    bottom = 5pt                        % 下边距
]
\fontsize{8.5pt}{10pt}\selectfont
You are a helpful assistant that can solve the given question step by step with the help of the wikipedia search tool and python interpreter tool. \
Given a question, you need to first think about the reasoning process in the mind and then provide the answer. \
During thinking, you can invoke the wikipedia search tool to search and python interpreter tool to calculate the math problem for fact information about specific topics if needed. \
The reasoning process and answer are enclosed within <think> </think> and <answer> </answer> tags respectively, \
and the search query and result are enclosed within <search> </search> and <result> </result> tags respectively. \
After receiving the search or python result, you should continue your reasoning process begin with <think>. \
For example, <think> This is the reasoning process. </think> <search> search query here </search> <result> search result here </result> \
<think> This is the reasoning process. </think> <python> python code here </python> <result> python interpreter result here </result> \
<think> This is the reasoning process. </think> <answer> The final answer is \[ \boxed{answer\ here} \] </answer>. \
In the last part of the answer, the final exact answer is enclosed within \textbackslash boxed\{\} with latex format.
\end{tcolorbox}

\subsection{Instructions for Code Debugger}
\label{app:instruction_debugger}

\begin{tcolorbox}[
    colframe=gray,       % Border color
    colback=gray!5!white,             % Background color
    coltitle=white,                   % Title font color
    coltext=black,                    % Text color
    fonttitle=\bfseries,              % Bold title font
    title=Instruction for Code Debugger,  % Title content
    boxrule=1pt,                      % Border width
    arc=2mm,                          % Rounded corners
    width=\linewidth,                 % Width
    left=7pt,                         % Left margin
    right=7pt,                        % Right margin,
    top=5pt,                          % Top margin,
    bottom=5pt                        % Bottom margin
]
\fontsize{8.5pt}{10pt}\selectfont
You are a code expert. I need you to debug the following code. Below are the code originally generated by the model and the error information that occurred during code execution. Please output ONLY the corrected Python code, without any explanation or markdown formatting:

**Inputs:**

**Original Code:**\\
\{code\}

**Execution Error:**\\
\{error\}

Output the corrected Python code only, without any explanation or markdown formatting:
\end{tcolorbox}

\subsection{Instructions for Reasoning Chain Refiner}
\label{app:instruction_refiner}

\begin{tcolorbox}[
    colframe=gray,       % Border color
    colback=gray!5!white,             % Background color
    coltitle=white,                   % Title font color
    coltext=black,                    % Text color
    fonttitle=\bfseries,              % Bold title font
    title=Instruction for Reasoning Chain Refiner,  % Title content
    boxrule=1pt,                      % Border width
    arc=2mm,                          % Rounded corners
    width=\linewidth,                 % Width
    left=7pt,                         % Left margin
    right=7pt,                        % Right margin,
    top=5pt,                          % Top margin,
    bottom=5pt                        % Bottom margin
]
\fontsize{8.5pt}{10pt}\selectfont
You are an expert in response refinement. Given a prompt and its corresponding response, your task is to compress and restructure the response by removing redundant, repetitive, or irrelevant content. Preserve all key information needed to directly and accurately address the original prompt. Only output your revised response and do not output anything else.\\
**Original Prompt:**\\
\{prompt\}\\
**Original Response:**\\
\{response\}\\
**Revised Response:**
\end{tcolorbox}

\subsection{Instructions for Standard RAG}
\label{app:instruction_standard_rag}

\begin{tcolorbox}[
    colframe=gray,       % Border color
    colback=gray!5!white,             % Background color
    coltitle=white,                   % Title font color
    coltext=black,                    % Text color
    fonttitle=\bfseries,              % Bold title font
    title=Instruction for Standard RAG,  % Title content
    boxrule=1pt,                      % Border width
    arc=2mm,                          % Rounded corners
    width=\linewidth,                 % Width
    left=7pt,                         % Left margin
    right=7pt,                        % Right margin,
    top=5pt,                          % Top margin,
    bottom=5pt                        % Bottom margin
]
\fontsize{8.5pt}{10pt}\selectfont
Find the useful content from the provided documents, then answer the question. Answer the question directly. Your response should be very concise. Please provide use 'So the final answer is:' as a prefix for the final answer. The following are given documents.

Documents: \{reference\}\\Question: \{question\}
\end{tcolorbox}

\subsection{Instructions for Knowledge-Intensive Reasoning Model}
\label{app:instruction_qa}

\begin{tcolorbox}[
    colframe=gray,       % Border color
    colback=gray!5!white,             % Background color
    coltitle=white,                   % Title font color,
    coltext=black,                    % Text color,
    fonttitle=\bfseries,              % Bold title font,
    title=Instruction for Sampling and Inferencing Knowledge-Intensive Datasets,  % Title content,
    boxrule=1pt,                      % Border width,
    arc=2mm,                          % Rounded corners,
    width=\linewidth,                 % Width,
    left=7pt,                         % Left margin,
    right=7pt,                        % Right margin,
    top=5pt,                          % Top margin,
    bottom=5pt                        % Bottom margin
]
\fontsize{8.5pt}{10pt}\selectfont
You are a helpful assistant that can solve the given question step by step with the help of the python interpreter tool. \
Given a question, you need to first think about the reasoning process in the mind and then provide the answer. \
During thinking, you can invoke the python interpreter tool to calculate the math problem for fact information about specific topics if needed. \
The reasoning process and answer are enclosed within <think> </think> and <answer> </answer> tags respectively. \
For example, <think> This is the reasoning process. </think> <python> python code here </python> <result> python interpreter result here </result> \
<think> This is the reasoning process. </think> <answer> The final answer is \[ \boxed{answer\ here} \] </answer>. \
In the last part of the answer, the final exact answer is enclosed within \textbackslash boxed\{\} with latex format.
\end{tcolorbox}

\subsection{Instructions for Computational Reasoning Model}
\label{app:instruction_math}

\subsubsection{Instructions for Direct Sampling and Reasoning Computational Datasets}
\label{app:instruction_openqa}

\begin{tcolorbox}[
    colframe=gray,       % Border color
    colback=gray!5!white,             % Background color
    coltitle=white,                   % Title font color,
    coltext=black,                    % Text color,
    fonttitle=\bfseries,              % Bold title font,
    title=Instruction for Sampling and Reasoning Computational Datasets,  % Title content,
    boxrule=1pt,                      % Border width,
    arc=2mm,                          % Rounded corners,
    width=\linewidth,                 % Width,
    left=7pt,                         % Left margin,
    right=7pt,                        % Right margin,
    top=5pt,                          % Top margin,
    bottom=5pt                        % Bottom margin
]
\fontsize{8.5pt}{10pt}\selectfont
A conversation between User and Assistant. The user asks a question, and the Assistant solves it. Please integrate natural language reasoning with programs to solve the problem above, and put your final answer within \textbackslash boxed\{\}.\\\{question\}
\end{tcolorbox}

\subsubsection{Instructions for Direct Sampling and Reasoning Knowledge-Intensive Datasets}
\label{app:instruction_math}

\begin{tcolorbox}[
    colframe=gray,       % Border color
    colback=gray!5!white,             % Background color
    coltitle=white,                   % Title font color
    coltext=black,                    % Text color
    fonttitle=\bfseries,              % Bold title font
    title=Instructions for Sampling and Reasoning Knowledge-Intensive Datasets,  % Title content
    boxrule=1pt,                      % Border width
    arc=2mm,                          % Rounded corners
    width=\linewidth,                 % Width
    left=7pt,                         % Left margin,
    right=7pt,                        % Right margin,
    top=5pt,                          % Top margin,
    bottom=5pt                        % Bottom margin
]
\fontsize{8.5pt}{10pt}\selectfont
A conversation between User and Assistant. The user asks a question, and the Assistant solves it. Please refer to the given document to solve the problem above, and put your final answer within \textbackslash boxed\{\}.\\Question: \{question\}\\Reference documents: \{reference\}
\end{tcolorbox}

\section{Case Study}
\label{app:case}

\subsection{TIR Cases of Tool-Star}

To better understand how Tool-Star performs tool-based collaborative reasoning, we conducted a case study to clearly illustrate its reasoning process. The model-generated think process are enclosed within \purple{<think>} and \purple{</think>}. The model-generated search queries are enclosed within \green{<search>} and \green{</search>}, and the model-generated Python codes are enclosed within \blue{<python>} and \blue{</python>}. Search results, Python execution results and final answers are enclosed within \red{<result>}, \red{</result>}, \red{<answer>} and \red{</answer>}. Our analysis reveals the following:

In terms of task difficulty, GAIA is complex deep web exploration tasks that typically involve both information retrieval and reasoning. When handling such tasks, Tool-Star often adopts a multi-tool collaboration strategy to complete the subtasks (as shown in the Table~\ref{tab:case_gaia}). In contrast, for domain-specific datasets like HotpotQA and AIME, which are more focused on knowledge retrieval or mathematical computation, Tool-Star can accurately identify the task type and select the most appropriate single tool for problem-solving (as shown in the Table~\ref{tab:case_hotpotqa} \& ~\ref{tab:case_aime}).

These observations demonstrate that Tool-Star can flexibly choose between tool combinations or single tools based on task requirements, showing strong task understanding and tool selection strategies. This highlights the effectiveness and autonomy of our approach in multi-tool collaborative reasoning.

\subsection{Cases of Self-Critic DPO Data}
\label{app:Cases of Self-Critic DPO Data}
To more comprehensively illustrate the data format used in our self-critic reinforcement learning framework, we provide several reward data example (positive \& negative pairs).

As shown in the Table~\ref{tab:case_1_n} to ~\ref{tab:case_4_n}, we can observe that the question and reasoning chain in the example are consistent with those in vanilla reinforcement learning data, ensuring no impact on model performance during subsequent training. The only difference in our approach lies in the addition of two key fields after the original reasoning chain: \textbf{Reward principle} and \textbf{Reward score}.

The Reward principle is automatically generated by a rule-based reward function, which identifies specific mismatches leading to the score, aiming to provide interpretability for the reward. For instance, in one case, a reward score of -1 is assigned because the number of special tokens in the model output does not align with the reference, resulting in a formatting inconsistency.

\begin{table*}[!t]
    \centering
    \caption{An example from Tool-Star on GAIA dataset, with special symbols used in think content, search queries, Python codes, returned results and final answer highlighted with \purple{purple box}, \green{green box}, \blue{blue box} and \red{red box}, respectively.}
    \fontsize{9pt}{11pt}\selectfont
    \begin{tabular}{p{0.98\linewidth}}
    \midrule
        \rowcolor{gray!20}\textbf{Example \#1 from GAIA Dataset} \\
    \midrule
        \textbf{Question:} \\
        The longest-lived vertebrate is named after an island.  According to Wikipedia, what is the 2023 population of that island, to the nearest thousand?\\
        \textbf{Labeled Answer:} \\
        56000 \\
    \midrule
        \rowcolor{gray!20}
        \textbf{Model Output by Tool Star}\\
    \midrule
        \purple{<think>} To find out which island this refers to and its corresponding population from 2020, I will start off by identifying the longest-lived vertebrate mentioned in Wikipedia.\purple{</think>}\\\green{<search>} search query: What is the longest-lived vertebrate according to Wikipedia?\green{</search>}\\\red{<result>}\\An international team of scientists led by the University of Copenhagen and including the University of Oxford has found that the Greenland shark has a life expectancy of at least 272 years. This discovery shows it is the longest living vertebrate known to science, exceeding even bowhead whales, turtles and tortoises.“Our results demonstrate that the Greenland shark is among the longest-lived vertebrate species surpassing even the bowhead whale (Balaena mysticetus, estimated longevity of 211 years),” Dr. Nielsen and co-authors said. “The life expectancy of the Greenland shark is exceeded only by that of the ocean quahog (Arctica islandica, 507 years).”\\\red{</result>} \\\purple{<think>} Now having identified the longest-lived vertebrate - the Greenland Shark, my next task would be finding out more details regarding the associated island specifically referred to in the context. Since the name \"Greenland\" directly connects to the island, let me proceed further with determining the population data related to this location.\purple{</think>}\\\green{<search>} search query: Population of Greenland as per the latest available statistics\green{</search>}\\\red{<result>}\\
        
        The current population of Greenland is 55,764 as of April 26, 2025, based on interpolation of the latest United Nations data. The population of Greenland is projected at 55,745, or 56k, as of July 1, 2025. The total population in Greenland is projected at 55,840, or 56k people, for the year 2024.  \textbf{（***More reasoning contentare omitted here***) }
        
        After the 1960´s, the population in the towns grew rapidly. As of January 1, 2023, the population of Greenland amounted to 56,609. The majority of the population held the Danish citizenship, as the country is an autonomous region of the Kingdom of...As of July 1, 2024 the population of Greenland reached 55,840, of which 29,360 people are male and 26,480 female.\\\red{</result>} \\\purple{<think>} Having gathered sufficient evidence through searching, it appears clear that the relevant piece of information comes from the statement mentioning the population of Greenland as of January 1, 2023 being approximately 56,609. Considering the requirement asks for rounding off to the nearest thousand, we should round this figure accordingly. Let's finalize the calculation and presentation of the result using Python code for accuracy.\purple{</think>}\\\blue{<python>}
        
        \begin{lstlisting}[style=python]
# Given data
population = 55840
# Rounding to the nearest thousand
rounded_population = round(population, -3)
print(rounded_population)
        \end{lstlisting}
        
        \blue{</python>}\\\red{<result>}\\56000\\\red{</result>}\\\purple{<think>}After performing the necessary calculations and ensuring alignment with the required precision level, the estimated population of Greenland to the nearest thousand is determined.\purple{</think>}\\\red{<answer>}\\\boxed{56000}\\\red{</answer>}\\
    \bottomrule
    \end{tabular}
    \label{tab:case_gaia}
\end{table*}

\begin{table*}[!t]
    \centering
    \caption{An example from Tool-Star on HotpotQA dataset, with special symbols used in think content, search queries, returned results and final answer highlighted with \purple{purple box}, \green{green box} and \red{red box}, respectively.}
    \fontsize{9pt}{11pt}\selectfont
    \begin{tabular}{p{0.98\linewidth}}
    \midrule
        \rowcolor{gray!20}\textbf{Example \#2 from HotpotQA Dataset} \\
    \midrule
        \textbf{Question:} \\
        D1NZ is a series based on what oversteering technique?\\
        \textbf{Labeled Answer:} \\
        Drifting \\
    \midrule
        \rowcolor{gray!20}
        \textbf{Model Output by Tool Star}\\
    \midrule
        \purple{<think>} Okay, so I need to figure out what D1NZ is a series based on an oversteering technique. Hmm, first off, I'm not too familiar with ``D1NZ.'' Let me start by breaking down the term. \\The name might be related to some racing or driving terminology. The acronym could stand for something like ``Driving 1 NZ'' or maybe it's a play on words. Oversteer is when the rear wheels lose traction, causing the car to slide outward. Common techniques include understeer (front losing grip) and oversteer. But how does this relate to a series?\\I should probably look up what D1NZ refers to. Maybe it's a motorsport series? Like rallycross or drag racing? Alternatively, perhaps it's a video game series involving cars where oversteer is a key mechanic. Wait, there was a game called ``Need for Speed: Most Wanted,'' but that's more about drifting. Or maybe ``Forza Horizon''? Not sure.\\Alternatively, ``D1NZ'' might be an abbreviation used in real-world racing categories. Let me try searching for ``D1NZ oversteering technique.'' That seems direct. \\Wait, another thought: sometimes series have names combining letters with numbers or other codes. Maybe D1 stands for something else. For example, in some contexts, D1 could refer to a division number. If it's a racing series, maybe D1 is a category, and NZ indicates New Zealand. So perhaps D1NZ is a New Zealand-based racing series focused on oversteer techniques.\\Let me structure my searches properly. First, check what D1NZ is. Search terms could be ``What is D1NZ racing series?'' or ``D1NZ oversteering technique.''\\So I'll start with the first search query:\purple{</think>}\\\green{<search>} D1NZ racing series oversteering technique \green{</search>}\\\red{<result>}\\``Drifting (motorsport)''\\Drifting (motorsport) Drifting is a driving technique where the driver intentionally oversteers, with loss of traction in the rear wheels or all tires, while maintaining control and driving the car through the entirety of a corner. Car drifting is caused when the rear slip angle is greater than the front slip angle, to such an extent that often the front wheels are pointing in the opposite direction to the turn (e.g. car is turning left, wheels are pointed right or vice versa, also known as opposite lock or counter-steering). As a motoring discipline, drifting competitions were first popularized in 1970s\\
        \textbf{（***More reasoning contentare omitted here***) }

        of Driftworks embarked on a journey to Japan where they would shoot a documentary about the local drifting scene, which would become known to the public as ``Outsiders''. Drifting (motorsport) Drifting is a driving technique where the driver intentionally oversteers, with loss of traction in the rear wheels or all tires, while maintaining control and driving the car through the entirety of a corner. Car drifting is caused when the rear slip angle is greater than the front slip angle, to such an extent that often the front wheels are pointing in the opposite direction to the turn (e.g. car\\\red{</result>} \\\purple{<think>} From the search results, it looks like D1NZ is a New Zealand drift racing series. The description mentions that it's a production car drifting series, focusing on oversteer techniques. Specifically, the explanation talks about intentional oversteer maneuvers where the driver loses traction in the rear wheels, allowing the car to spin or slide around corners. This aligns with the concept of drift racing, which relies heavily on oversteer to execute controlled slides and spins. Therefore, the oversteering technique associated with D1NZ must involve intentionally inducing rear-wheel loss of traction to achieve controlled drifts during races or competitions.\purple{</think>}\\\red{<answer>}\\The final answer is \boxed{drifting}\\\red{</answer>}\\
    \bottomrule
    \end{tabular}
    \label{tab:case_hotpotqa}
\end{table*}

\begin{table*}[!t]
    \centering
    \caption{An example from Tool-Star on AIME24 dataset.}
    \fontsize{9pt}{11pt}\selectfont
    \begin{tabular}{p{0.98\linewidth}}
    \midrule
        \rowcolor{gray!20}\textbf{Example \#3 from AIME24 Dataset} \\
    \midrule
        \textbf{Question:} \\
        Every morning Aya goes for a $9$-kilometer-long walk and stops at a coffee shop afterwards. When she walks at a constant speed of $s$ kilometers per hour, the walk takes her 4 hours, including $t$ minutes spent in the coffee shop. When she walks $s+2$ kilometers per hour, the walk takes her 2 hours and 24 minutes, including $t$ minutes spent in the coffee shop. Suppose Aya walks at $s+\frac{1}{2}$ kilometers per hour. Find the number of minutes the walk takes her, including the $t$ minutes spent in the coffee shop.\\
        \textbf{Labeled Answer:} \\
        204 \\
    \midrule
        \rowcolor{gray!20}
        \textbf{Model Output by Tool Star}\\
    \midrule

        \purple{<think>} To solve this problem, we need to set up a system of equations based on the information provided. Let's break down the problem step-by-step.\\1. **Define Variables:**\\   - \( s \): Aya's walking speed in kilometers per hour.\\   - \( t \): Time spent in the coffee shop in minutes (which needs to be converted to hours since the other times are in hours).\\   - Distance walked each time: 9 kilometers.\\2. **Formulate Equations Based on Given Information:**\\   - When Aya walks at \( s \) km/h, the total time taken is 4 hours, which includes \( t \) minutes spent in the coffee shop. Converting \( t \) minutes to hours, we get \( \frac{t}{60} \). So the equation becomes:    \[     \frac{9}{s} + \frac{t}{60} = 4     \]\\   - When Aya walks at \( s + 2 \) km/h, the total time taken is 2 hours and 24 minutes, which is \( 2 + \frac{24}{60} = 2.4 \) hours. Converting \( t \) minutes to hours again, we get \( \frac{t}{60} \). So the equation becomes:     \[     \frac{9}{s + 2} + \frac{t}{60} = 2.4     \]3. **Solve the System of Equations:**\\   We now have two equations:   \[  \begin{cases}   \frac{9}{s} + \frac{t}{60} = 4 \\   \frac{9}{s + 2} + \frac{t}{60} = 2.4\\   \end{cases}   \]Let's use Python to solve these equations.\purple{</think>}\\\blue{<python>}

        \begin{lstlisting}[style=python]
from sympy import symbols, Eq, solve
# Define the variables
s, t = symbols('s t')
# Define the equations based on the given information
eq1 = Eq(9/s + t/60, 4)
eq2 = Eq(9/(s + 2) + t/60, 2.4)
# Solve the system of equations
solution = solve((eq1, eq2), (s, t))
print(solution)
        \end{lstlisting}
        
        \blue{</python>}\\\red{<result>}\[(-4.50000000000000, 360.000000000000), (2.50000000000000, 24.0000000000000)\]\red{</result>} \\\purple{<think>} 
        
        % The solutions to the system of equations are \((-4.5, 360)\) and \((2.5, 24)\). Since a negative walking speed does not make sense in this context, we discard \((-4.5, 360)\) as an invalid solution. Therefore, the valid solution is \(s = 2.5\) km/h and \(t = 24\) minutes.\\
        \textbf{（***More reasoning contentare omitted here***) }
        % Now, we need to find out how long the walk takes when Aya walks at \(s + \frac{1}{2}\) km/h. First, let's calculate \(s + \frac{1}{2}\):\[s + \frac{1}{2} = 2.5 + 0.5 = 3 \text{ km/h}\]The time taken to walk 9 kilometers at 3 km/h is:\[\text{Time} = \frac{9}{3} = 3 \text{ hours}\]Since \(t = 24\) minutes, the total time including the \(t\) minutes spent in the coffee shop is:\[3 \text{ hours} + \frac{24}{60} \text{ hours} = 3 + 0.4 = 3.4 \text{ hours}\]Converting 3.4 hours back to minutes:\[3.4 \times 60 = 204 \text{ minutes}\]

        \purple{</think>}\\\red{<answer>}\\The final answer is \boxed{204}\\\red{</answer>}\\
    \bottomrule
    \end{tabular}
    \label{tab:case_aime}
\end{table*}

\begin{table*}[!t]
    \centering
    \caption{\textbf{Example 1 with positive output} of our self-critic reward finetuning dataset, including the question, model output, corresponding reward principle and reward score. The special symbols used in think content, Python codes, returned results and final answer highlighted with \purple{purple box}, \blue{blue box} and \red{red box}, respectively.}
    \fontsize{9pt}{11pt}\selectfont
    \begin{tabular}{p{0.98\linewidth}}
    \midrule
        \rowcolor{gray!20}\textbf{Example \#1 from Training Dataset} \\
    \midrule
        \textbf{Question:} \\

Find the sum $ \frac{1}{1 \cdot 3} + \frac{1}{3 \cdot 5} + \dots + \frac{1}{(2n - 1)(2n + 1)} + \dots + \frac{1}{255 \cdot 257} $. The original answer is in $ \frac{k}{m} $ format, where the fraction is in its simplest form. Please give the value of $ k + m $.\\
 \textbf{Labeled Answer:} \\
        385 \\

    \midrule
        \rowcolor{gray!20}
        \textbf{Positive Output of Tool-Star}\\
    \midrule

\purple{<think>}Okay, so I need to find the sum of this series: $1/(1 \cdot 3) + 1/(3 \cdot 5) + \dots + 1/(255 \cdot 257)$. Hmm, the general term seems to be $1/[(2n - 1)(2n + 1)]$ and we are summing up to some $n$ where $2n - 1 = 255$. Let me figure out how many terms there are first. If the last term is $1/(255 \cdot 257)$, then $2n - 1 = 255$, so solving for $n$:\purple{</think>}  \\
\blue{<python>}
\begin{lstlisting}[style=python]
n = (255 + 1) // 2
print(n)
\end{lstlisting}
\blue{</python>}  \\
\red{<result>}
128
\red{</result>}  \\
\purple{<think>}So there are 128 terms in total. Now, I remember that for fractions like $1/[(2n - 1)(2n + 1)]$, partial fractions might be helpful here. Let me try to decompose the general term. Suppose that:

$ \frac{1}{(2n - 1)(2n + 1)} = \frac{A}{2n - 1} + \frac{B}{2n + 1} $

Multiplying both sides by $(2n - 1)(2n + 1)$ gives:

$ 1 = A(2n + 1) + B(2n - 1) $

% Now, let's solve for $A$ and $B$ by expanding:

% $ 1 = (2A + 2B)n + (A - B) $

% So, equating coefficients: $2A + 2B = 0$ and $A - B = 1$. From the first equation, $A = -B$. Substituting into the second gives $-B - B = 1 \Rightarrow -2B = 1 \Rightarrow B = -1/2$, so $A = 1/2$.

\textbf{***More reasoning contents are omitted here***}
% Therefore, the decomposition is:

% $ \frac{1}{(2n - 1)(2n + 1)} = \frac{1}{2} \left( \frac{1}{2n - 1} - \frac{1}{2n + 1} \right) $

% So each term can be written as half the difference of reciprocals of consecutive odd numbers. When we sum them up, there might be telescoping.

% Let me write out the first few terms to see:

% For $n = 1$: $\frac{1}{2}(1/1 - 1/3)$  
% For $n = 2$: $\frac{1}{2}(1/3 - 1/5)$  
% For $n = 3$: $\frac{1}{2}(1/5 - 1/7)$  
% ...  
% For $n = 128$: $\frac{1}{2}(1/255 - 1/257)$

% So when we add all these up, the intermediate terms should cancel out. The sum becomes:

% $ \frac{1}{2} \left[ (1/1 - 1/3) + (1/3 - 1/5) + (1/5 - 1/7) + \dots + (1/255 - 1/257) \right] $

% Looking at the terms inside the brackets: $1/1 - 1/3 + 1/3 - 1/5 + 1/5 - 1/7 + \dots + 1/255 - 1/257$, everything cancels except the first term $1/1$ and the last term $-1/257$. So the sum becomes:\purple{</think>}  \\
\blue{<python>}
\begin{lstlisting}[style=python]
total = (1 - 1/257) / 2
numerator = total.numerator
denominator = total.denominator
print(f"{numerator}/{denominator}")
\end{lstlisting}
\blue{</python>}\\  
\red{<result>}
128/257
\red{</result>}\\  
\purple{<think>}So the sum is $128/257$. Let me check if this can be simplified.\purple{</think>}\\
\blue{<python>}
\begin{lstlisting}[style=python]
from fractions import Fraction
frac = Fraction(128, 257)
print(f"Simplified: {frac.numerator}/{frac.denominator}")
\end{lstlisting}
\blue{</python>} \\
\red{<result>}
Simplified: 128/257
\red{</result>}\\  
\purple{<think>}Since 128 and 257 share no common divisors other than 1, the fraction is already in simplest form. Therefore, the answer is $128/257$, and $k + m = 128 + 257 = 385$.\purple{</think>}\\  
\red{<answer>}
\boxed{385}
\red{</answer>}
\\
    
    \bottomrule
        \rowcolor{gray!20}
        \textbf{Reward Principle of Output}\\
    \midrule

The response format is correct. The final answer is correct. The reasoning chain contains single tool usage.
\\
    \bottomrule
        \rowcolor{gray!20}
        \textbf{Reward Score of Output}\\
    \midrule
    \textcolor{deepred}{So the reward score is: 1} \\

    \bottomrule
    \end{tabular}
    \label{tab:case_1_p}
\end{table*}

\begin{table*}[!t]
    \centering
    \caption{\textbf{Example 1 with negative output} of our self-critic reward finetuning dataset, including the question, model output, corresponding reward principle and reward score. The special symbols used in think content, Python codes, returned results and final answer highlighted with \purple{purple box}, \blue{blue box} and \red{red box}, respectively.}
    \fontsize{9pt}{11pt}\selectfont
    \begin{tabular}{p{0.98\linewidth}}
    \midrule
        \rowcolor{gray!20}\textbf{Example \#1 from Training Dataset} \\
    \midrule
        \textbf{Question:} \\

Find the sum $ \frac{1}{1 \cdot 3} + \frac{1}{3 \cdot 5} + \dots + \frac{1}{(2n - 1)(2n + 1)} + \dots + \frac{1}{255 \cdot 257} $. The original answer is in $ \frac{k}{m} $ format, where the fraction is in its simplest form. Please give the value of $ k + m $.\\
 \textbf{Labeled Answer:} \\
        385 \\

    \midrule
        \rowcolor{gray!20}
        \textbf{Positive Output of Tool-Star}\\
    \midrule

\purple{<think>}Okay, so I have this series to compute: $\frac{1}{1 \cdot 3} + \frac{1}{3 \cdot 5} + \dots + \frac{1}{(2n-1)(2n+1)} + \dots + \frac{1}{255 \cdot 257}$. And the problem says that the answer should be in the form $\frac{k}{m}$, which is already simplified, and we need to find $k + m$. Alright, let's start breaking this down. First, notice that each term in the series has the denominator as a product of two consecutive odd numbers. Specifically, the general term seems like $\frac{1}{(2n-1)(2n+1)}$. Maybe there's a way to simplify this term using partial fractions?

Let me recall: when dealing with denominators like $(a - b)(a + b)$, we can use partial fractions to split it into simpler terms. So, $\frac{1}{(2n-1)(2n+1)} = \frac{A}{2n-1} + \frac{B}{2n+1}$. To find $A$ and $B$, I'll combine the right-hand side over a common denominator: $\frac{A(2n+1) + B(2n-1)}{(2n-1)(2n+1)} = \frac{1}{(2n-1)(2n+1)}$. So the numerator must equal 1: $A(2n + 1) + B(2n - 1) = 1$. Expanding this gives: $2An + A + 2Bn - B = 1$, which simplifies to $(2A + 2B)n + (A - B) = 1$.

For this equation to hold true for all $n$, the coefficients of $n$ and the constant term must both match on both sides. Therefore, we get the system of equations: (1) $2A + 2B = 0$, and (2) $A - B = 1$. From equation (1), we get $A = -B$. Substituting into equation (2): $-B - B = 1$ implies $-2B = 1$, so $B = -\frac{1}{2}$ and $A = \frac{1}{2}$.

Therefore, the partial fraction decomposition is: $\frac{1}{(2n-1)(2n+1)} = \frac{1}{2} \left( \frac{1}{2n-1} - \frac{1}{2n+1} \right)$. This is much cleaner! Each term can be rewritten as half the difference between reciprocals of adjacent odd numbers.

Now, the entire series becomes: $\sum_{n=1}^{128} \frac{1}{(2n-1)(2n+1)} = \frac{1}{2} \sum_{n=1}^{128} \left( \frac{1}{2n-1} - \frac{1}{2n+1} \right)$. This looks like a telescoping series! When expanded, many terms will cancel out.

Let’s verify: when written out, the series is $\frac{1}{1 \cdot 3} + \frac{1}{3 \cdot 5} + \cdots + \frac{1}{255 \cdot 257}$, and using the telescoping identity, this becomes:\purple{</think>}

\blue{<python>}
\begin{lstlisting}[style=python]
# Calculate the sum using telescoping series formula
total_terms = 128
term_start = 1
term_end = 257

# Compute the telescoping sum
telescoping_sum = (1 / term_start) - (1 / term_end)
print(f"Telescoping sum: {telescoping_sum}")
\end{lstlisting}
\blue{</python>}  

\red{<result>}
Telescoping sum: 0.9961089494163424
\red{</result>}

\purple{<think>}The result from the code execution shows the telescoping sum converges to approximately $0.9961089494163424$, which is very close to $\frac{256}{257}$. Thus, the sum simplifies to $\frac{256}{257}$. Checking the code confirms this approximation is accurate enough, considering floating-point precision limitations. Finally, since 256 and 257 are coprime (their greatest common divisor is 1), the fraction $\frac{256}{257}$ is already in its simplest form. Hence, $k = 256$ and $m = 257$, leading to $k + m = 513$.\purple{</think>}

\red{<answer>}
\boxed{513}
\red{</answer>}

\\
    
    \bottomrule
        \rowcolor{gray!20}
        \textbf{Reward Principle of Output}\\
    \midrule

The response format is correct. The answer is incorrect.
\\
    \bottomrule
        \rowcolor{gray!20}
        \textbf{Reward Score of Output}\\
    \midrule
    \textcolor{deepred}{So the reward score is: 0} \\

    \bottomrule
    \end{tabular}
    \label{tab:case_1_n}
\end{table*}

\begin{table*}[!t]
    \centering
    \caption{\textbf{Example 2 with positive output} of our self-critic reward finetuning dataset, including the question, model output, corresponding reward principle and reward score. The special symbols used in think content, Python codes, returned results and final answer highlighted with \purple{purple box}, \blue{blue box} and \red{red box}, respectively.}
    \fontsize{9pt}{11pt}\selectfont
    \begin{tabular}{p{0.98\linewidth}}
    \midrule
        \rowcolor{gray!20}\textbf{Example \#1 from Training Dataset} \\
    \midrule
        \textbf{Question:} \\

How many four-digit positive integers are multiples of 7?\\
 \textbf{Labeled Answer:} \\
        1286 \\

    \midrule
        \rowcolor{gray!20}
        \textbf{Positive Output of Tool-Star}\\
    \midrule
\purple{<think>}  
To determine how many four-digit positive integers are multiples of 7, we consider all integers from 1000 to 9999 and find those divisible by 7. The smallest four-digit number is 1000. Since \(1000 \div 7 \approx 142.857\), the smallest multiple of 7 greater than or equal to 1000 is \(143 \times 7 = 1001\). Similarly, the largest four-digit number is 9999. Since \(9999 \div 7 \approx 1428.42857\), the largest multiple of 7 less than or equal to 9999 is \(1428 \times 7 = 9996\). These numbers form an arithmetic sequence starting at 1001 and ending at 9996, with a common difference of 7. The total number of terms is given by the formula \(n = \frac{a_n - a_1}{d} + 1\), where \(a_1 = 1001\), \(a_n = 9996\), and \(d = 7\). Substituting, we get \(n = \frac{9996 - 1001}{7} + 1 = \frac{8995}{7} + 1 = 1285 + 1 = 1286\).  
\purple{</think>}  

\blue{<python>} 
\begin{lstlisting}[style=python]
# Calculate the smallest four-digit multiple of 7  
smallest_multiple = (1000 + 6) // 7 * 7  

# Calculate the largest four-digit multiple of 7  
largest_multiple = (9999 // 7) * 7  

# Calculate the number of four-digit multiples of 7  
number_of_multiples = (largest_multiple - smallest_multiple) // 7 + 1  
print(number_of_multiples)\end{lstlisting}
\blue{</python>}  

\red{<result>}
1286  
\red{</result>}  

\red{<answer>}  
The number of four-digit positive integers that are multiples of 7 is \(\boxed{1286}\).  
\red{</answer>}  
\\
    
    \bottomrule
        \rowcolor{gray!20}
        \textbf{Reward Principle of Output}\\
    \midrule

The response format is correct. The final answer is correct. The reasoning chain contains single tool usage.
\\
    \bottomrule
        \rowcolor{gray!20}
        \textbf{Reward Score of Output}\\
    \midrule
    \textcolor{deepred}{So the reward score is: 1} \\

    \bottomrule
    \end{tabular}
    \label{tab:case_2_p}
\end{table*}

\begin{table*}[!t]
    \centering
    \caption{\textbf{Example 2 with negative output} of our self-critic reward finetuning dataset, including the question, model output, corresponding reward principle and reward score. The special symbols used in think content, Python codes, returned results and final answer highlighted with \purple{purple box}, \blue{blue box} and \red{red box}, respectively.}
    \fontsize{9pt}{11pt}\selectfont
    \begin{tabular}{p{0.98\linewidth}}
    \midrule
        \rowcolor{gray!20}\textbf{Example \#1 from Training Dataset} \\
    \midrule
        \textbf{Question:} \\

How many four-digit positive integers are multiples of 7?\\
 \textbf{Labeled Answer:} \\
        1286 \\

    \midrule
        \rowcolor{gray!20}
        \textbf{Positive Output of Tool-Star}\\
    \midrule
\purple{<think>}  
Okay, so I need to figure out how many four-digit positive integers are multiples of 7. A four-digit number ranges from 1000 to 9999. The smallest four-digit multiple of 7 should be 1001, since \(1000 \div 7 \approx 142.857\), and rounding up gives \(143 \times 7 = 1001\). The largest one would be 9996, since \(9999 \div 7 \approx 1428.428\), and rounding down gives \(1428 \times 7 = 9996\). So, we have the range starting at 1001 and ending at 9996 inclusive. Now, the problem reduces to finding the number of terms in an arithmetic sequence where the first term is 1001, the common difference is 7, and the last term is 9996. The formula for the nth term of an arithmetic sequence is \(a_n = a_1 + (n - 1)d\), where \(a_n\) is the last term, \(a_1\) is the first term, \(d\) is the common difference, and \(n\) is the number of terms. Solving for \(n\), we get \(n = \frac{a_n - a_1}{d} + 1\). Substituting the values \(a_n = 9996\), \(a_1 = 1001\), and \(d = 7\), we get \(n = \frac{9996 - 1001}{7} + 1 = \frac{8995}{7} + 1 = 1285 + 1 = 1286\).  
\purple{</think>}  
\blue{<python>} 
\blue{<python>} 
\begin{lstlisting}[style=python]
# Calculate the number of terms in the arithmetic sequence  
first_term = 1001  
last_term = 9998  
common_difference = 7  

n = ((last_term - first_term) // common_difference) + 1  
print(n) 
\end{lstlisting}
\blue{</python>}  

\red{<result>}  
1286  
\red{</result>}  

\red{<answer>}  
The calculation shows there are 1286 four-digit multiples of 7 between 1000 and 9999.  
\red{</answer>}
\\
    
    \bottomrule
        \rowcolor{gray!20}
        \textbf{Reward Principle of Output}\\
    \midrule

The response format is incorrect, <python> and </python> are not matched. 
\\
    \bottomrule
        \rowcolor{gray!20}
        \textbf{Reward Score of Output}\\
    \midrule
    \textcolor{deepred}{So the reward score is: -1} \\

    \bottomrule
    \end{tabular}
    \label{tab:case_2_n}
\end{table*}

\begin{table*}[!t]
    \centering
    \caption{\textbf{Example 3 with positive output} of our self-critic reward finetuning dataset, including the question, model output, corresponding reward principle and reward score. The special symbols used in think content, search queries, returned results and final answer highlighted with \purple{purple box}, \green{blue box} and \red{red box}, respectively.}
    \fontsize{9pt}{11pt}\selectfont
    \begin{tabular}{p{0.98\linewidth}}
    \midrule
        \rowcolor{gray!20}\textbf{Example \#1 from Training Dataset} \\
    \midrule
        \textbf{Question:} \\

The main symptoms of overtraining syndrome in athletes include which of the following?\\A. Improved Fitness\\B. Mental Clarity\\C. Muscle Strength\\D. Weight Gain\\
 \textbf{Labeled Answer:} \\
        D \\

    \midrule
        \rowcolor{gray!20}
        \textbf{Positive Output of Tool-Star}\\
    \midrule
\purple{<think>}Okay, so the question is about the main symptoms of overtraining syndrome in athletes, and the options are A through D. Overtraining syndrome occurs when an athlete trains excessively without adequate recovery, leading to negative effects on performance and health. The key is to identify which of the given options are consistent with known symptoms. Option A is Improved Fitness, which contradicts the nature of overtraining, since performance typically declines. Option B is Mental Clarity, but overtraining is associated with mental fatigue, irritability, and poor concentration, so mental clarity is unlikely. Option C is Muscle Strength, but chronic overtraining impairs recovery and can lead to decreased strength rather than improvement. Option D is Weight Gain, which at first seems counterintuitive, as overtraining might lead to weight loss due to increased energy expenditure. However, hormonal imbalances and stress responses could result in fluid retention or metabolic changes that cause weight gain. To confirm, I should search for the main symptoms of overtraining syndrome.\purple{</think>}  
\green{<search>}overtraining syndrome main symptoms\green{</search>}  

\red{<result>}Overtraining syndrome main symptoms include: fatigue, persistent muscle soreness, unexplained weight loss or gain, increased susceptibility to illness and injury, gastrointestinal issues, sleep disturbances, elevated resting heart rate, decreased maximal heart rate during exercise, psychological symptoms like anxiety, irritability, depression, burnout, mood swings, emotional exhaustion, and performance decline. These symptoms worsen with inadequate recovery and may require medical evaluation.\red{</result>}  

\purple{<think>}Based on the search result, overtraining syndrome includes both physical and psychological symptoms. Among the physical symptoms, unexplained weight gain is explicitly mentioned, making Option D valid. Option A (Improved Fitness) contradicts the performance decline seen in overtraining. Option B (Mental Clarity) is inconsistent with the psychological symptoms like irritability and burnout. Option C (Muscle Strength) is unlikely, as overtraining typically leads to decreased strength due to insufficient recovery. Since weight gain is a listed symptom, and the others are not, the correct answer is D.\purple{</think>}  

\red{<answer>}\(\boxed{D}\)\red{</answer>} 
\\
    
    \bottomrule
        \rowcolor{gray!20}
        \textbf{Reward Principle of Output}\\
    \midrule

The response format is correct. The final answer is correct. The reasoning chain contains single tool usage.
\\
    \bottomrule
        \rowcolor{gray!20}
        \textbf{Reward Score of Output}\\
    \midrule
    \textcolor{deepred}{So the reward score is: 1} \\

    \bottomrule
    \end{tabular}
    \label{tab:case_3_p}
\end{table*}

\begin{table*}[!t]
    \centering
    \caption{\textbf{Example 3 with negative output} of our self-critic reward finetuning dataset, including the question, model output, corresponding reward principle and reward score. The special symbols used in think content, search queries, returned results and final answer highlighted with \purple{purple box}, \green{blue box} and \red{red box}, respectively.}
    \fontsize{9pt}{11pt}\selectfont
    \begin{tabular}{p{0.98\linewidth}}
    \midrule
        \rowcolor{gray!20}\textbf{Example \#1 from Training Dataset} \\
    \midrule
        \textbf{Question:} \\

The main symptoms of overtraining syndrome in athletes include which of the following?\\A. Improved Fitness\\B. Mental Clarity\\C. Muscle Strength\\D. Weight Gain\\
 \textbf{Labeled Answer:} \\
        D \\

    \midrule
        \rowcolor{gray!20}
        \textbf{Positive Output of Tool-Star}\\
    \midrule
\purple{<think>}In this case study, we are tasked with identifying the main symptom of overtraining syndrome among four options: A through D. Overtraining syndrome occurs when an athlete engages in excessive training without adequate recovery, leading to a decline in performance and various physical and psychological issues. Let's analyze each option:

Option A: "Improved Fitness" — This is generally a result of balanced training and recovery. However, overtraining leads to a plateau or decline in fitness, not improvement. Therefore, improved fitness is not a symptom of overtraining.

Option B: "Mental Clarity" — Overtraining is associated with mood disturbances, irritability, anxiety, and difficulty concentrating. These symptoms reflect a decline in mental clarity, making "mental clarity" an unlikely symptom.

Option C: "Muscle Strength" — While proper training enhances muscle strength, overtraining without sufficient recovery can lead to muscle breakdown and decreased strength. Thus, increased muscle strength is not a symptom of overtraining.

Option D: "Weight Gain" — This is more nuanced. Overtraining can sometimes lead to hormonal imbalances, particularly elevated cortisol, which may result in fluid retention or increased fat storage. Additionally, some athletes may overeat in response to chronic fatigue or stress, potentially leading to weight gain. While not the most common symptom, weight gain can occur in certain cases.

\textbf{***More reasoning contents are omitted here, over 15K tokens***}

To verify, key symptoms of overtraining syndrome include: persistent fatigue, decreased performance, mood changes (e.g., depression, irritability), sleep disturbances, elevated resting heart rate, frequent illnesses, and hormonal imbalances. None of the options perfectly match these core symptoms, but among the given choices, weight gain (D) is the most plausible indirect symptom, while the others describe positive outcomes that are contrary to overtraining effects.

Therefore, based on the evidence and reasoning, the best answer is:\purple{</think>}\\\red{<answer>}\boxed{D}\red{</answer>}

\\
    
    \bottomrule
        \rowcolor{gray!20}
        \textbf{Reward Principle of Output}\\
    \midrule

The response format is incorrect, the response over max length.
\\
    \bottomrule
        \rowcolor{gray!20}
        \textbf{Reward Score of Output}\\
    \midrule
    \textcolor{deepred}{So the reward score is: -1} \\

    \bottomrule
    \end{tabular}
    \label{tab:case_3_n}
\end{table*}

\begin{table*}[!t]
    \centering
    \caption{\textbf{Example 4 with positive output} of our self-critic reward finetuning dataset, including the question, model output, corresponding reward principle and reward score. The special symbols used in think content, search queries, Python codes, returned results and final answer highlighted with \purple{purple box}, \green{green box}, \blue{blue box} and \red{red box}, respectively.}
    \fontsize{9pt}{11pt}\selectfont
    \begin{tabular}{p{0.98\linewidth}}
    \midrule
        \rowcolor{gray!20}\textbf{Example \#1 from Training Dataset} \\
    \midrule
        \textbf{Question:} \\

We now define an algorithm:

The definition of $a(n)$ is:  
The nearest integer to the imaginary part of the $n$-th nontrivial zero of the Riemann zeta function, where the zeros are listed in order of increasing positive imaginary parts.

Given the input list:
$x_{\text{list}} = [50,\ 51,\ 52,\ 53,\ 54,\ 55,\ 56,\ 57,\ 58,\ 59]$

Determine the corresponding output sequence $y_{\text{list}}$. Which of the following options is correct?

Option A:$
[145,\ 148,\ 149,\ 152,\ 153,\ 155,\ 158,\ 160,\ 161,\ 163]$

Option B:$
[143,\ 146,\ 147,\ 150,\ 151,\ 153,\ 156,\ 158,\ 159,\ 161]$

Option C:$
[147,\ 150,\ 151,\ 154,\ 155,\ 157,\ 160,\ 162,\ 163,\ 165]$

Option D:$
[139,\ 142,\ 143,\ 146,\ 147,\ 149,\ 152,\ 154,\ 155,\ 157]
$

\\
 \textbf{Labeled Answer:} \\
        B \\

    \midrule
        \rowcolor{gray!20}
        \textbf{Positive Output of Tool-Star}\\
    \midrule
\purple{<think>}To find the correct sequence, we need to know the nearest integers to the imaginary parts of the 50th through 59th nontrivial zeros of the Riemann zeta function. I will search for this information or use a reliable source to find these values.\purple{</think>}

\green{<search>}nearest integer to the imaginary part of the 50th through 59th nontrivial zeros of the Riemann zeta function\green{</search>}\red{<result>}\red{</result>}\purple{<think>}It seems like the direct search did not yield a straightforward list of the required values. To get the exact values, I will use a Python library that can compute the zeros of the Riemann zeta function, such as mpmath. I will use the mpmath library to compute the 50th through 59th nontrivial zeros of the Riemann zeta function and extract the nearest integers to their imaginary parts.\purple{</think>}

\blue{<python>}
\begin{lstlisting}[style=python]
import mpmath

# Set the precision
mpmath.mp.dps = 50

# Function to get the nearest integer to the imaginary part of the n-th zero
def nearest_integer_imaginary_part(n):
    zero = mpmath.zetazero(n)
    return round(zero.imag)

# Input list
x_list = [50, 51, 52, 53, 54, 55, 56, 57, 58, 59]

# Compute the output sequence
y_list = [nearest_integer_imaginary_part(n) for n in x_list]

# Display the result
y_list
\end{lstlisting}

\blue{</python>}

\red{<result>}[143, 146, 147, 150, 151, 153, 156, 158, 159, 161]\red{</result>}

\purple{<think>}The computed sequence of the nearest integers to the imaginary parts of the 50th through 59th nontrivial zeros of the Riemann zeta function is [143, 146, 147, 150, 151, 153, 156, 158, 159, 161]. This matches option B.\purple{</think>}\red{<answer>}The correct answer is \(\boxed{B}\).\red{</answer>}
\\
    
    \bottomrule
        \rowcolor{gray!20}
        \textbf{Reward Principle of Output}\\
    \midrule

The response format is correct. The final answer is correct. The reasoning chain contains multiple tool usage.
\\
    \bottomrule
        \rowcolor{gray!20}
        \textbf{Reward Score of Output}\\
    \midrule
    \textcolor{deepred}{So the reward score is: 1.1} \\

    \bottomrule
    \end{tabular}
    \label{tab:case_4_p}
\end{table*}

\begin{table*}[!t]
    \centering
    \caption{\textbf{Example 4 with negative output} of our self-critic reward finetuning dataset, including the question, model output, corresponding reward principle and reward score. The special symbols used in think content, Python codes, returned results and final answer highlighted with \purple{purple box}, \blue{blue box} and \red{red box}, respectively.}
    \fontsize{9pt}{11pt}\selectfont
    \begin{tabular}{p{0.98\linewidth}}
    \midrule
        \rowcolor{gray!20}\textbf{Example \#1 from Training Dataset} \\
    \midrule
        \textbf{Question:} \\

We now define an algorithm:

The definition of $a(n)$ is:  
The nearest integer to the imaginary part of the $n$-th nontrivial zero of the Riemann zeta function, where the zeros are listed in order of increasing positive imaginary parts.

Given the input list:
$x_{\text{list}} = [50,\ 51,\ 52,\ 53,\ 54,\ 55,\ 56,\ 57,\ 58,\ 59]$

Determine the corresponding output sequence $y_{\text{list}}$. Which of the following options is correct?

Option A:$
[145,\ 148,\ 149,\ 152,\ 153,\ 155,\ 158,\ 160,\ 161,\ 163]$

Option B:$
[143,\ 146,\ 147,\ 150,\ 151,\ 153,\ 156,\ 158,\ 159,\ 161]$

Option C:$
[147,\ 150,\ 151,\ 154,\ 155,\ 157,\ 160,\ 162,\ 163,\ 165]$

Option D:$
[139,\ 142,\ 143,\ 146,\ 147,\ 149,\ 152,\ 154,\ 155,\ 157]
$

\\
 \textbf{Labeled Answer:} \\
        B \\

    \midrule
        \rowcolor{gray!20}
        \textbf{Positive Output of Tool-Star}\\
    \midrule
\purple{<think>}To solve this, I need to compute the imaginary parts of the 50th through 59th nontrivial zeros of the Riemann zeta function and round them to the nearest integers. I will use the mpmath library to perform these computations.\purple{</think>}

\blue{<python>}
\begin{lstlisting}[style=python]
import mpmath

# Set high precision for accurate computation
mpmath.mp.dps = 50  # Decimal places of precision

# Function to get the nearest integer to the imaginary part of the n-th nontrivial zero of the Riemann zeta function
def nearest_integer_imag_part(n):
    zero = mpmath.zetazero(n)
    return round(zero.imag)

# List of indices for which to compute the zeros
x_list = list(range(50, 60))

# Compute the nearest integer imaginary parts for the given indices
y_list = [nearest_integer_imag_part(n) for n in x_list]

# Output the result
print(y_list)
\end{lstlisting}
\blue{</python>}

\red{<result>}[143, 146, 147, 150, 151, 153, 156, 158, 159, 161]\red{</result>}

\purple{<think>}The computed values are [143, 146, 147, 150, 151, 153, 156, 158, 159, 161]. However, due to a misreading of the result, I will now incorrectly match this to option A instead of the correct option B.\purple{</think>}

The correct answer is \boxed{A}.
\\
    
    \bottomrule
        \rowcolor{gray!20}
        \textbf{Reward Principle of Output}\\
    \midrule

The response format is incorrect. <answer> and </answer> are not matched.
\\
    \bottomrule
        \rowcolor{gray!20}
        \textbf{Reward Score of Output}\\
    \midrule
    \textcolor{deepred}{So the reward score is: -1} \\

    \bottomrule
    \end{tabular}
    \label{tab:case_4_n}
\end{table*}

\section{Limitations and Future Directions}
\label{app:Limitations}

Despite our best efforts to optimize the overall Tool-Star pipeline, there remain certain limitations and areas for improvement that warrant further exploration.
\begin{itemize}[leftmargin=1em]
\item \textbf{Expanding Tool Diversity:} Currently, the Tool-Star framework integrates three tools during training and three tools during inference. Compared to prior work that relies on a single tool, our approach represents a pioneering and timely attempt. However, given the growing diversity of available tools, we believe many more remain to be explored for enhancing reasoning capabilities. For instance, incorporating vision-language models (VLMs) as external tools could further unlock visual understanding, and adopting the Model Context Protocol may enable more flexible tool invocation. We plan to continue expanding the types and integration of tools in future work, advancing multi-tool collaborative reasoning.

\item \textbf{Scaling Backbone Model Parameters:} Due to limited computational resources and the time-consuming rollout process in multi-tool reasoning, this work focuses on models with 0.5B, 1.5B, and 3B parameters for RL training and evaluation. Nonetheless, our “Scaling Analysis on Parameter Scales” (see Table~\ref{tab:toolstar_rl}) provides preliminary evidence of the Tool-Star framework’s scalability. In future work, we aim to extend our experiments to larger models, such as 7B and 32B, to assess generalization across more complex tasks and model capacities.
\end{itemize}
\textbf{Outlook:} Achieving reliable and robust step-by-step reasoning remains a challenging task that demands continued research and innovation. We sincerely hope to collaborate with researchers in the fields of alignment and reasoning to jointly address these shared challenges and further advance tool-integrated reasoning systems.

\section{Broader Impacts}
\label{app:Broader}
This study presents the Tool-Star framework, which empowers large language models with multi-tool collaborative reasoning, enhancing their adaptability and effectiveness in complex tasks. Through reinforcement learning and scalable data synthesis, Tool-Star offers a generalizable and flexible reasoning paradigm, demonstrating strong performance across benchmarks and broad potential in domains such as education, science, and engineering.

However, as models gain greater autonomy in tool use, concerns arise regarding inappropriate tool selection, opaque reasoning, and bias propagation—especially in high-stakes scenarios. To ensure responsible deployment, future development should prioritize controllability, interpretability, and robust oversight, aligning technological progress with ethical standards and societal values.

\end{CJK}
\end{document}